\pdfoutput=1

\documentclass[10pt,journal,compsoc]{IEEEtran}

\usepackage[nocompress]{cite}
\usepackage[pdftex]{graphicx}
\graphicspath{{Figures/}}
\usepackage{amsmath}
\usepackage{amssymb}
\usepackage{url}
\usepackage{multirow}
\usepackage{xcolor}

\begin{document}

\title{Towards Protecting Face Embeddings\\in Mobile Face Verification Scenarios}

\author{Vedrana~Krivoku\'ca~Hahn
        and~S\'ebastien~Marcel \\
        \textit{Idiap Research Institute (Martigny, Switzerland)}  
}

\IEEEtitleabstractindextext{%
\begin{abstract}
This paper proposes PolyProtect, a method for protecting the sensitive face embeddings that are used to represent people's faces in neural-network-based face verification systems. PolyProtect transforms a face embedding to a more secure template, using a mapping based on multivariate polynomials parameterised by user-specific coefficients and exponents. In this work, PolyProtect is evaluated on two open-source face recognition systems in a cooperative-user mobile face verification context, under the toughest threat model that assumes a fully-informed attacker with complete knowledge of the system and all its parameters. Results indicate that PolyProtect can be tuned to achieve a satisfactory trade-off between the recognition accuracy of the PolyProtected face verification system and the irreversibility of the PolyProtected templates. Furthermore, PolyProtected templates are shown to be effectively unlinkable, especially if the user-specific parameters employed in the PolyProtect mapping are selected in a non-naive manner. The evaluation is conducted using practical methodologies with tangible results, to present realistic insight into the method's robustness as a face embedding protection scheme in practice. This work is fully reproducible using the publicly available code at: \url{https://gitlab.idiap.ch/bob/bob.paper.polyprotect_2021}. 
\end{abstract}

}

\maketitle

\section{Introduction}

The reliance on automated face recognition systems to authenticate a person's identity is becoming commonplace. Whether we are attempting to unlock our smartphones or cross a country border, success (or failure) is increasingly defined by a machine's ability to recognise the link between our faces and our identities. Despite the numerous security benefits associated with this form of authentication, there are growing privacy concerns over how our sensitive face data is being handled by the increasing number of applications that are demanding this personal information for identity management. 

Modern face recognition systems are based on deep learning architectures, whereby a compact face representation (commonly referred to as an \textit{embedding}) is learned from several example images of a person's face. Recently, it has been shown that a face embedding can be inverted to recover an approximation of the original face image \cite{z16, c17, m19}, and that certain soft biometric attributes (e.g., sex, race, age, hair colour) can be extracted from the representative face embeddings \cite{f20, t20}. These findings indicate that face embeddings contain a wealth of personally identifiable information, which, if leaked from the system(s) in which they are employed, would represent a threat to the privacy of face recognition system users. 

To ensure the public's trust in face recognition technologies, it is imperative that we establish effective means of protecting the employed face embeddings. This is especially important in light of the recent EU General Data Protection Regulation (GDPR)\footnote{\url{https://bit.ly/3nMM1Qz}}, which imposes a legal obligation to exercise caution in handling biometric data to protect individuals' digital identities. Considering the urgency of this matter in view of the widespread use of face recognition in practice, work investigating the protection of face embeddings is surprisingly limited. 

This paper aims to contribute towards preserving the privacy of face recognition system users, by proposing a method for converting our sensitive face embeddings to more secure representations. In particular, we propose a new method, called \textit{PolyProtect}, which maps face embeddings to protected templates with the help of multivariate polynomials, whose parameters are defined separately for each user of the underlying face recognition system. Considering the proliferating use of face recognition in mobile devices, such as smartphones, we present a comprehensive evaluation of PolyProtect in a cooperative-user mobile face verification scenario, on two open-source face recognition systems. Our results indicate that PolyProtect shows promise as an effective face embedding protection scheme in practice.   

The remainder of the paper is structured as follows. Section 2 outlines the main approaches to face embedding protection in the literature; Section 3 proposes our new protection method, PolyProtect; Section 4 analyses PolyProtect in terms of the three most common evaluation criteria; and Section 5 presents conclusions and avenues for future work.

\section{Face embedding protection methods}

It is generally agreed upon that a face embedding protection method should possess the following three properties:

\begin{enumerate}

	\item \textbf{Recognition accuracy:} The incorporation of the protection method into a face recognition system should not result in a (significant) degradation of the system's recognition accuracy. 
	
	\item \textbf{Irreversibility:} It should be impossible (or computationally infeasible) to recover the original face embedding (or face image) from its protected version.
	
	\item \textbf{Unlinkability:} It should be possible to generate multiple, sufficiently different protected templates from the same subject's face embeddings, such that the templates cannot be linked to the same identity. This would allow for the renewal (replacement) of compromised templates and the use of the same face identity across multiple applications, without the risk of cross-matching the protected templates.
	
\end{enumerate}

In the literature, there are two main types of approaches towards face embedding protection, as illustrated in Fig. 1. The first type of approach consists of learning face embeddings from face images using a neural network optimised for this task, then mapping the learned embeddings to protected templates using a \textit{separate, handcrafted} protection algorithm. The second type of approach involves training a neural network to \textit{learn} a suitable protection algorithm, to transform a face image to its protected template. 

\begin{figure}[!ht]
\centering
\includegraphics[width=\columnwidth]{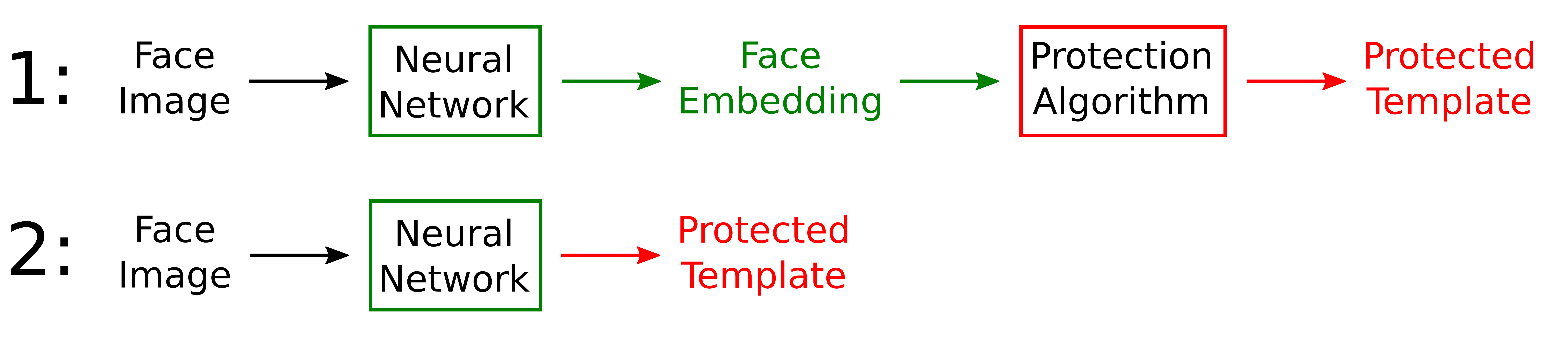}%
\caption{The two main types of approaches to face embedding protection.}
\end{figure}

Examples of the \textit{first} type of approach to face embedding protection, include: \cite{p15, a20, a19, s19, r21, p19, b18}. These methods use some sort of algorithmically defined transformation to convert the face embeddings (learned from the input face images) to more secure representations. The proposed transformations include one-way cryptographic \cite{p15} or Winner Takes All \cite{a20} hashing, convolution of the embedding with a random kernel \cite{a19}, use of the Fuzzy Commitment \cite{s19} or Fuzzy Vault \cite{r21} scheme, fusion of a subject's face embedding with a different subject's face embedding using keys extracted from the two sets of features \cite{p19}, and homomorphic encryption \cite{b18}. The main issue with these approaches is that they have not been comprehensively evaluated in terms of their ability to simultaneously satisfy \textit{all three} properties of face embedding protection methods. More specifically, although an evaluation of the recognition accuracy is presented for all the methods, the irreversibility and unlinkability analyses often lack depth. In particular, the irreversibility tends to be either: (i) assumed (e.g., based on the secrecy of certain transformation parameters or on the reputed irreversibility of the employed transforms) but not empirically justified in the evaluation context, or (ii) estimated from a purely theoretical point of view that does not reflect the method's robustness to an inversion attack in practice (where certain theoretical assumptions are unlikely to hold). Similarly, the renewability of protected templates is usually simply assumed by virtue of the randomness of external parameters, but the unlinkability property is seldom experimentally validated. So, we do not have a complete picture of each method's strengths and weaknesses. 

Examples of the \textit{second} type of approach to face embedding protection, include: \cite{p16, c18, k18, c19, t19, m20, p21, l21}. These methods \textit{learn} the protected template from the input face image (with face embeddings being extracted, in some format, during the process). In \cite{p16, c18, k18, c19}, a random code is \textit{pre-defined} for each subject during enrollment, then the neural network is trained to map different samples of the same subject's face to their (same) corresponding code. A cryptographic hash of the random code represents the protected face template. The main limitation of these approaches is that, since the protected templates are pre-defined, we would need to re-train the neural network for the enrollment of each new user or the re-enrollment of existing users whose protected templates have been compromised. To get around this problem, the methods in \cite{t19, m20, p21, l21} train a neural network to learn its \textit{own representation} of a protected template, instead of training it to learn a mapping to a pre-defined (pre-hash) code. In \cite{t19}, the neural network is trained to map face images to intermediate binary codes then correct errors in these binary codes (followed by cryptographic hashing of the error-corrected codes). Since the network is trained to learn the same code for every presentation of the same subject's face, renewal of compromised protected templates would be impossible. To enable template renewability, \cite{m20, p21, l21} train their neural networks to incorporate external, user-specific randomness into the process of learning the protected templates. These methods appear promising in their ability to generate renewable and unlinkable protected templates, but there is no analysis providing insight into the expected scalability of the renewability effort before neural network re-training may need to be invoked. Furthermore, the irreversibility of most of the methods in the \textit{type 2} protection approach has been evaluated in terms of the \textit{final protected template}, based on certain assumptions (e.g., the secrecy of user-specific parameters or the one-way property of cryptographic hash functions); however, there has been no consideration of how irreversibility may be affected by potential information leakage in different layers of the trained neural network, assuming a fully-informed attacker with access to the network and its learned parameters. So, we have insufficient evidence of the methods' irreversibility in practice, particularly for a fully-informed attacker. 

Considering both types of approaches towards face embedding protection (Fig. 1), the \textit{second} type appears to be gaining traction recently. This is because the promise of designing a potentially complex protection algorithm without the need to explicitly define it, is attractive. Unfortunately, this approach has several limitations. Firstly, the protection method is specific to the neural network within which it has been trained, so it cannot be readily adopted for the protection of face embeddings generated by other face recognition models. Secondly, template renewability appears infeasible or impractical for all but methods such as \cite{m20, p21, l21}; however, more empirical evidence is needed to justify the scalability of these methods without invoking network re-training. Thirdly, when we train a neural network to \textit{learn} a protection algorithm for the input face images, there is often uncertainty about what exactly the network is learning at each stage of the process, which makes it difficult to perform a comprehensive evaluation of the irreversibility of the resulting protected templates. This is because, if we assume the most challenging threat scenario where an adversary has access to the trained model (i.e., network architecture and all learned parameters), then the irreversibility analysis should consider how this knowledge could be used to extract additional information about the original face embedding or image from different layers of the neural network. Thus far, none of the face embedding protection methods in the literature has presented such a thorough irreversibility analysis, so we do not have a fair picture of how irreversible the protected face templates would be in practice.

For the aforementioned reasons, our work focuses on the \textit{first} type of approach, where a handcrafted protection algorithm is applied to learned face embeddings. This approach allows for: (1) a flexible protection method that can be applied to face embeddings generated using different face recognition models; (2) the enrollment of new users or re-enrollment of compromised users without the need to re-train any neural network; and (3) a more precise definition of the protection algorithm (without the uncertainty in neural-network-based learning), and thus a better understanding of appropriate evaluation techniques. Although a few face embedding protection methods in this category have already been proposed (e.g., \cite{p15, a20, a19, s19, r21, p19, b18}), they are incomplete in their evaluations (as explained earlier), making it difficult to draw concrete conclusions on their expected robustness in practice.

In light of this discussion, the main contribution of this paper is a new protection method, PolyProtect, which transforms face embeddings to their protected counterparts via user-specific multivariate polynomials. The following two sections describe PolyProtect and present an evaluation of its suitability as a face embedding protection method. It should be emphasized that our focus was on evaluating PolyProtect from a \textit{practical} point of view, particularly when analysing its irreversibility, which is usually neglected in favour of theoretical approaches in the literature. The presented results are, therefore, more tangible and realistic than the idealistic outcomes of purely theoretical evaluations, which makes it easier to grasp PolyProtect's practical value. Although theoretical evaluations can be valuable in certain cases, we would nevertheless encourage other researchers to also consider adopting practical methodologies when evaluating their proposed protection methods. This would help to provide a clearer picture of the methods' robustness in practice, thereby allowing for more direct method comparisons in specific application contexts.

\section{PolyProtect}

This section proposes PolyProtect, a new method for protecting face embeddings in neural-network-based face recognition systems. Let $V = [v_1, v_2, ..., v_n]$ denote an \textit{n}-dimensional, real-number face embedding. The aim of PolyProtect is to map $V$ to another real-number feature vector, $P = [p_1, p_2, ..., p_k]$ (where $k < n$), which is the protected version of $V$.  This is achieved by mapping sets of $m$ (where $m << n$) consecutive elements from $V$ to single elements in $P$ via multivariate polynomials defined by a set of $m$ user-specific (i.e., distinct for each user of the face recognition system), ordered, unique, non-zero integer coefficients, $C = [c_1, c_2, ..., c_m]$, and exponents, $E = [e_1, e_2, ..., e_m]$. 

The first $m$ consecutive elements of $V$ (i.e., $v_1, v_2, ..., v_m$) are mapped to the first element in $P$ (i.e., $p_1$) via Eq. (1):

\begin{equation}
p_1 = c_{1}v_{1}^{e_1} + c_{2}v_{2}^{e_2} + ... + c_{m}v_{m}^{e_m}
\end{equation}

The elements of $V$ used to generate $p_2$ depend on the desired amount of \textit{overlap} between successive sets of elements. The \textit{minimum} overlap is 0, in which case the elements of $V$ in each set would be unique. The \textit{maximum} overlap is $m - 1$, in which case successive element sets would share $m - 1$ elements. Eqs. (2) and (3) define the mapping from $V$ to $p_2$ for overlaps of 0 and $m - 1$, respectively:

\begin{equation}
p_2 = c_{1}v_{m+1}^{e_1} + c_{2}v_{m+2}^{e_2} + ... + c_{m}v_{m+m}^{e_m}	
\end{equation}

\begin{equation}
p_2 = c_{1}v_{2}^{e_1} + c_{2}v_{3}^{e_2} + ... + c_{m}v_{m+1}^{e_m}	
\end{equation}

The remaining elements in $P$ (i.e., $p_3, ..., p_k$) are generated in a similar manner, until all the elements in $V$ have been used up. If the last set of elements from $V$ is incomplete because the dimensionality of $V$ is not divisible by the required number of element sets (defined by $m$ and the amount of overlap), $V$ is padded by a sufficient number of zeros to complete the last set. Fig. 2 illustrates the mapping from $V$ to $P$ for overlaps of 0 to 4, when $V$ consists of 128 elements and $m = 5$.

\begin{figure*}
\centering
\includegraphics[width=\textwidth]{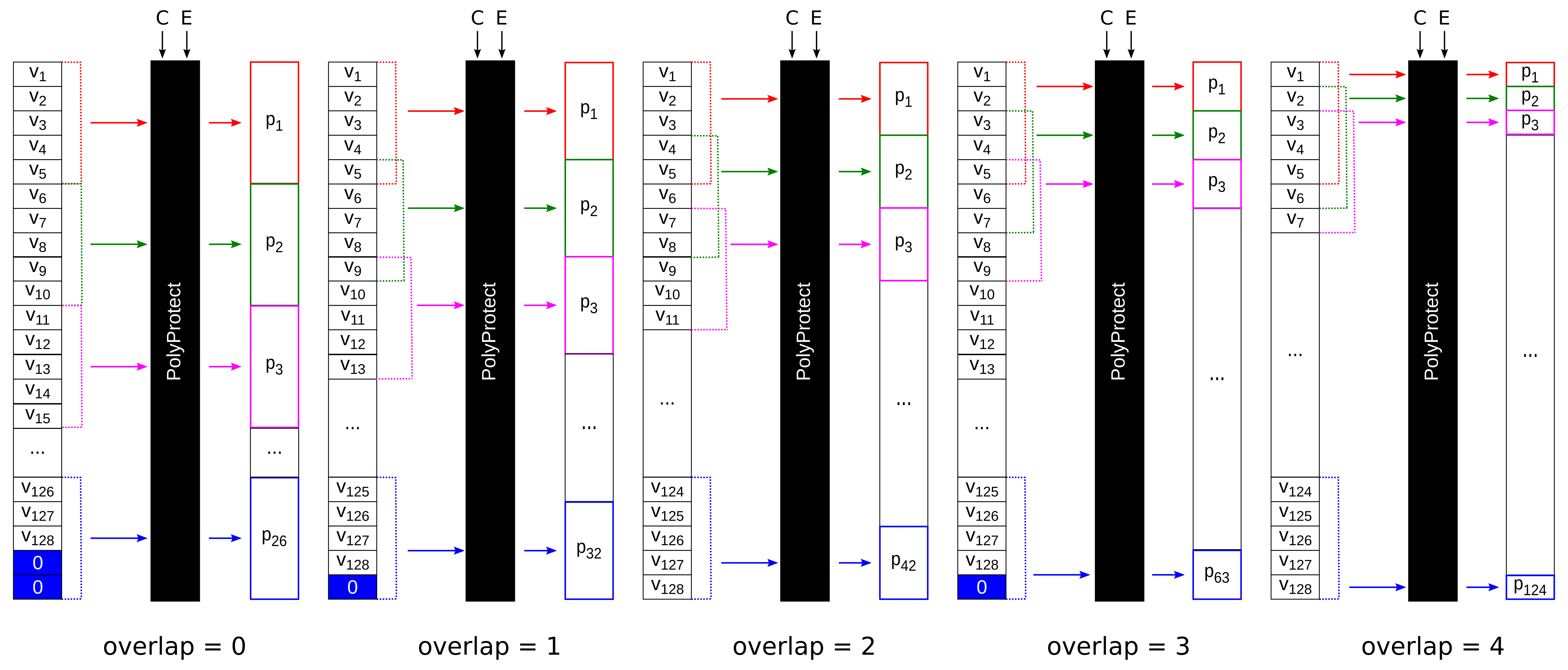}%
\caption{Mapping 128-dimensional $V$ to $P$ via PolyProtect, using $C = [c_1, c_2, ..., c_5]$ and $E = [e_1, e_2, ..., e_5]$, for different amounts of overlap.}
\end{figure*}

From Fig. 2, it is evident that the dimensionality of $P$ is influenced by the amount of overlap used in the $V \rightarrow P$ mapping. Using an overlap of 0 results in the \textit{smallest} $P$ (consisting of 26 elements), and using an overlap of 4 results in the \textit{largest} $P$ (consisting of 124 elements). So, it is reasonable to conclude that, the greater the amount of overlap, the more information from $V$ will be contained in $P$. This will be shown to have an effect on the recognition accuracy and irreversibility properties of PolyProtect, which are evaluated in Section 4. 

The main idea behind PolyProtect was to design a protection algorithm that introduces user-specific, tuneable non-linearities to a face embedding, such that the resulting protected template would be irreversible even if all the parameters of the mapping are known. The use of multiple multivariate polynomials with user-specific coefficients and exponents thus seemed like a natural choice. The requirement that the mapping from a face embedding to its protected template be user-specific, was motivated by a desire to satisfy the unlinkability property (see Section 1). Furthermore, we wished to be able to easily tune the protection algorithm to control the trade-off between the recognition accuracy and irreversibility properties, which is achievable in PolyProtect primarily by varying the amount of overlap.  Since PolyProtect relies on \textit{user-specific} parameters ($C$ and $E$), its envisioned operating scenario is in cooperative-user face verification applications (such as proving one's identity in order to unlock a personal smartphone). 

To the best of our knowledge, PolyProtect represents a novel approach towards face embedding protection. Although the reader may be tempted to liken PolyProtect to the Fuzzy Vault scheme \cite{js06} due to the use of polynomials in both methods, there are actually a number of fundamental differences between the two approaches. Firstly, the Fuzzy Vault scheme operates on \textit{unordered} sets of elements, whereas PolyProtect relies on an \textit{ordered} feature vector (e.g., an embedding). Secondly, the polynomial used in the Fuzzy Vault scheme is \textit{univariate} (i.e., each element in the biometric template serves as the input to the polynomial in turn), whereas PolyProtect's polynomials are \textit{multivariate} (i.e., groups of elements from the biometric template are simultaneously passed to the polynomial). Thirdly, the Fuzzy Vault scheme relies on the addition of random data points (\textit{chaff points}) to hide the polynomial outputs, whereas this is \textit{not} a feature of PolyProtect. Fourthly, the verification operation in the Fuzzy Vault scheme requires the reconstruction of the secret polynomial to extract its coefficients, which serve as the user's secret \textit{key}. The verification of PolyProtected templates does \textit{not} require polynomial reconstruction nor the extraction of a secret key; instead, PolyProtected templates are compared directly using a distance/similarity metric. So, we may conclude that PolyProtect and the Fuzzy Vault scheme represent fundamentally different approaches to biometric template protection.

Although its construction suggests the suitability of PolyProtect for protecting \textit{any} real-number biometric feature vector, the focus of this paper is on its applicability to \textit{face embeddings} alone.  Similarly, although PolyProtect is envisioned for adoption in cooperative-user verification scenarios in general, the evaluation presented in this paper will target only a \textit{mobile application context}. This is because: (i) facial recognition in mobile devices (e.g., smartphones) is one of the most (if not \textit{the} most) common face verification applications in practice, and (ii) this type of facial recognition represents the least constrained (and thus most challenging) cooperative-user face verification scenario. The suitability of PolyProtect to serve as an effective face embedding protection scheme in this scenario in practice, is evaluated in Section 4.

\section{Analysis of PolyProtect for face\\ embeddings in mobile face verification}

This section evaluates the suitability of PolyProtect for securing face embeddings in a cooperative-user mobile face verification scenario. Section 4.1 describes our experimental set-up. Then, Sections 4.2 to 4.4, respectively, evaluate PolyProtect in terms of the three properties outlined in Section 1: recognition accuracy, irreversibility, and unlinkability.

\subsection{Experimental set-up}

To evaluate the suitability of PolyProtect for protecting face embeddings, we first needed to establish a \textit{baseline} deep-neural-network-based face recognition system, into which PolyProtect could be incorporated. We adopted \textit{two} open-source systems for this purpose, both implemented within the \texttt{bob.bio.face\_ongoing} PyPI package\footnote{\url{https://bit.ly/2XLsYLQ}}: \textit{facenet} and \textit{idiap\_msceleb\_inception\_v2\_centerloss\_rgb}, henceforth referred to as Facenet and Idiap, respectively. The main difference between the two systems lies in how they generate face embeddings from face images, which is defined by their adopted pre-trained deep neural network models: Facenet uses the open-source FaceNet model 20170512-110547 from David Sandberg\footnote{\url{https://bit.ly/39oYNMV}}, and Idiap uses a CNN model based on the Inception-ResNet-v2 architecture and trained on the MS-Celeb-1M dataset\footnote{\url{https://bit.ly/3Alwivn}}. Both the Facenet and Idiap systems work with 128-dimensional face embeddings, which were the inputs to our PolyProtect algorithm. 

As noted earlier, we focus on face \textit{verification} rather than \textit{identification}. Fig. 3 illustrates the incorporation of PolyProtect into the enrollment and verification stages of our two baseline systems, which are differentiated by their feature extractors (i.e., deep neural network models trained to extract 128-dimensional face embeddings from face images).

\begin{figure*}[!h]
\centering
\includegraphics[width=\textwidth]{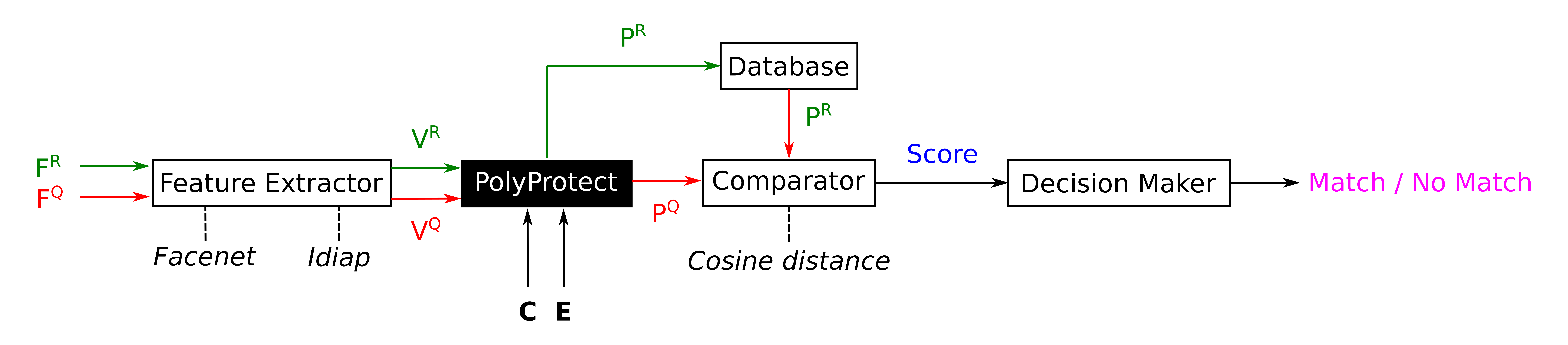}%
\caption{Enrollment (green arrows) and verification (red arrows) in the PolyProtected Facenet and Idiap face recognition systems. $F^R$ and $F^Q$ are the reference and query input face images, $V^R$ and $V^Q$ are the corresponding 128-dimensional face embeddings, and $P^R$ and $P^Q$ are the PolyProtected templates. The PolyProtect mapping is defined by the user-specific coefficients, $C$, and exponents, $E$.}
\end{figure*}  

Fig. 3 indicates that, during enrollment, the reference face embedding, $V^R$, is protected using PolyProtect (defined by the user-specific parameters $C$ and $E$), then the resulting protected template, $P^R$, is stored in the system's database. During verification, the query face embedding, $V^Q$, is likewise protected to generate $P^Q$. Then, $P^Q$ and $P^R$ are compared using the cosine distance metric, and the resulting score is processed to determine whether or not the underlying faces match. The cosine distance metric was chosen because the same metric was used to compare the unprotected face embeddings in our baseline systems (i.e., $V^R$ and $V^Q$), which makes it easier to evaluate the effect of PolyProtect on the systems' recognition accuracy. Technically, however, the score output by the Comparator in Fig. 3 is a \textit{similarity} score, because the cosine distances are multiplied by -1 (to turn them into similarity scores). So, the score range is [-2.0, 0.0], where -2.0 would indicate that $P^R$ and $P^Q$ (or $V^R$ and $V^Q$) are as different as possible, while 0.0 would imply that they are the same.

Note that the reason for selecting the Facenet and Idiap face recognition systems was that they were reported\footnote{\url{https://bit.ly/39jfCIT}} to have the best recognition accuracy when evaluated in the verification scenario on the face dataset that we deemed the most suitable for evaluating PolyProtect: Mobio \cite{m12}, which consists of bi-modal (audio and video) data captured from 152 people using two mobile devices (a phone and a laptop) in a cooperative-user scenario. We chose to evaluate PolyProtect on this dataset for four main reasons. Firstly, recall that the aim of this paper is to evaluate the suitability of PolyProtect for protecting face embeddings in a cooperative-user mobile face verification context. This is because the prevalence of mobile devices in our society, as well as current market trends, suggests that this is likely to be one of the (if not \textit{the}) biggest uses for face verification in practice (e.g., unlocking your smartphone with your face), so this represents PolyProtect's most likely application scenario. Since the Mobio database most closely imitates this target scenario, it was a very fitting choice for the PolyProtect evaluations presented in this paper. Secondly, the face videos in Mobio were captured in uncontrolled environments, so the face images extracted from the video frames are realistic and natural in terms of illumination, head poses, and facial expressions. Thirdly, the samples were acquired over 2 years across 5 countries, and 12 sessions in total were captured per person (database subject). This, together with the uncontrolled acquisition environment, makes the database challenging in the amount of session variability it exhibits. Fourthly, the dataset is publicly available\footnote{\url{https://bit.ly/39n3pDi}}, which allows for the reproducibility of our experiments.

To evaluate the PolyProtected systems, we first needed to establish three parameters: the value of $m$, and the ranges of $C$ and $E$. Recall that $m$ specifies the number of elements from the face embedding, $V$, used to generate each element in the protected template, $P$. We chose $m = 5$, meaning that each element in $P$ was generated using 5 consecutive elements from $V$, as illustrated for different overlaps in Fig. 2. Our thinking behind setting $m = 5$ was inspired by the Abel-Ruffini theorem, which states that there is no closed-form algebraic expression for solving polynomials of degree 5 or higher with arbitrary coefficients \cite{g18}. While this does not imply that it is impossible to find the roots of such polynomials, what it means is that a general expression does not exist (unlike, for example, for 2-degree polynomials). So, an attacker trying to reverse the $V \rightarrow P$ mapping to recover $V$ could not rely on an analytical approach using an existing, well-defined formula. While they could attempt to use a root-finding algorithm to find a numerical approximation for $V$, such methods are generally sensitive to initial guesses and are, therefore, prone to converging to a false solution. (More details on the feasibility of recovering $V$ from $P$ are provided in the irreversibility analysis in Section 4.3.) The reason we did not set $m > 5$ was because this would require using exponents larger than 5 in the PolyProtect mapping. Since the face embeddings consist of quite small floating point values, large powers would effectively obliterate certain embedding elements during the PolyProtect mapping. In the same vein of thought, the exponents, $E$, in our PolyProtected systems were randomly-permuted, unique integers in the range [1, 5]. 

The choice of a suitable range for the coefficients, $C$, was not as evident. We experimented with several $C$ ranges to generate the PolyProtected templates, but there appeared to be no significant differences in the resulting recognition accuracy of our PolyProtected face verification systems. We suspect that this is due to the use of the cosine distance metric in the comparison of PolyProtected templates. Since this metric calculates the difference between the \textit{directions} of the vectors being compared, their magnitudes are less important, so the effects of using a larger or smaller $C$ range are presumably diluted as a result. We expect that employing a magnitude-sensitive metric (e.g., Euclidean distance) would result in larger differences in the recognition accuracy of the PolyProtected systems when different $C$ ranges are employed; however, at this stage we have chosen to adopt the cosine distance metric for consistency with the comparison of unprotected face embeddings in the baseline systems, and we leave the investigation of alternative metrics to future work. For the evaluations presented in this paper, we used an arbitrarily selected $C$ range of [-50, 50], so all sets of $C$s consisted of 5 randomly-selected, unique, non-zero integers in this range.

We are now ready to present our evaluation of PolyProtect, when the protection method is incorporated into the Facenet and Idiap face verification systems, and when the analysis is performed on the Mobio dataset. The evaluation is based on the three properties of protection methods outlined in Section 1: recognition accuracy, irreversibility, and unlinkability. Sections 4.2 to 4.4, respectively, present the corresponding analysis. As mentioned in Section 2, our focus was on evaluating PolyProtect from a \textit{practical} point of view (as opposed to theoretical), to present a clearer picture of the method's practical value.

\subsection{Recognition accuracy}

This section investigates how the recognition accuracy of the PolyProtected  face verification systems compares to that of the corresponding baseline (unprotected) systems. The aim of this analysis was to determine whether the incorporation of PolyProtect into a deep-neural-network-based face verification system would degrade its recognition accuracy. 

To conduct this analysis, the recognition accuracy of each baseline system (Facenet and Idiap) was first evaluated on the Mobio dataset, by running the \textit{mobile0-male} verification protocol\footnote{\url{https://bit.ly/3nP9HnF}} used to generate the reported baseline results\footnote{\url{https://bit.ly/39jfCIT}}. The same protocol was then applied to the corresponding PolyProtected face verification systems.

The recognition accuracy of our PolyProtected face verification systems was
evaluated in two scenarios: (i) Normal (best-case), and (ii) Stolen Coefficients and Exponents (worst-case). The PolyProtected system should operate in the Normal scenario most of the time. Here, we assume that each enrolled user dutifully employs their own $C$ and $E$ parameters in the generation of their PolyProtected face templates, as envisioned by the design of the PolyProtect scheme. In the Stolen Coefficients and Exponents scenario, a user attempts to pass off as a different user by stealing the target's $C$ and $E$ parameters, and applying them to their own face embedding to generate their PolyProtected template. While it is reasonable to assume that the latter scenario should be uncommon in practice (provided that users' $C$ and $E$ parameters are stored securely), it is still important to consider this worst-case scenario (as is sometimes done in the literature) when analysing the expected recognition accuracy of PolyProtect in practice.

The Normal (N) scenario was simulated by randomly generating a set of \textit{different} $C$ and $E$ parameters for \textit{each subject}, then applying those parameters to map each of the subject's reference and query embeddings to their corresponding PolyProtected templates. So, for reference and query face embeddings from the \textit{same} subject, the \textit{same} $C$ and $E$ parameters were used to generate their corresponding PolyProtected templates, while reference and query face embeddings from \textit{different} subjects were protected using \textit{different} $C$ and $E$ parameters (since these parameters are subject/user-specific). Next, comparison scores were computed between all required pairs of reference and query PolyProtected templates (as defined by the aforementioned verification protocol), resulting in a set of \textit{genuine} scores (when the templates being compared originate from the \textit{same} subject) and a set of \textit{impostor} scores (when the templates originate from \textit{different} subjects). 

To simulate the Stolen Coefficients and Exponents (SCE) scenario, we assumed the most extreme scenario where each subject's $C$ and $E$ parameters are stolen and used by the other subjects. So, for all query face embeddings that were meant to be compared to a particular reference embedding (according to the adopted verification protocol), the \textit{same} $C$ and $E$ parameters (i.e., those belonging to the reference identity) were used to generate \textit{all} the corresponding query PolyProtected templates. Consequently, although the genuine scores were calculated in the same way for both the N and SCE scenarios, the impostor scores were calculated using \textit{subject}-specific parameters in the N scenario and \textit{reference}-specific parameters in the SCE scenario.

This process was repeated for 10 trials, where for each trial a new set of $C$ and $E$ parameters was chosen for each subject. (This may be interpreted as simulating 10 different applications in which the same subjects are enrolled.) Then, the genuine and impostor scores across the 10 trials were concatenated (separately), and the recognition accuracy was calculated on the concatenated scores.  These same sets of $C$ and $E$ parameters (10 per subject) were applied to each PolyProtected system (Facenet and Idiap), as well as for each \textit{overlap} parameter defining the PolyProtect mapping. Fig. 4 depicts the resulting ROC plots, generated on the \textit{evaluation} set of the Mobio database, in the N and SCE scenarios. Each plot compares the verification accuracy of the corresponding baseline system against the verification accuracy of the PolyProtected system for different amounts of overlap used in the PolyProtect mapping.  

\begin{figure}[!h]
\centering
\includegraphics[width=0.49\columnwidth]{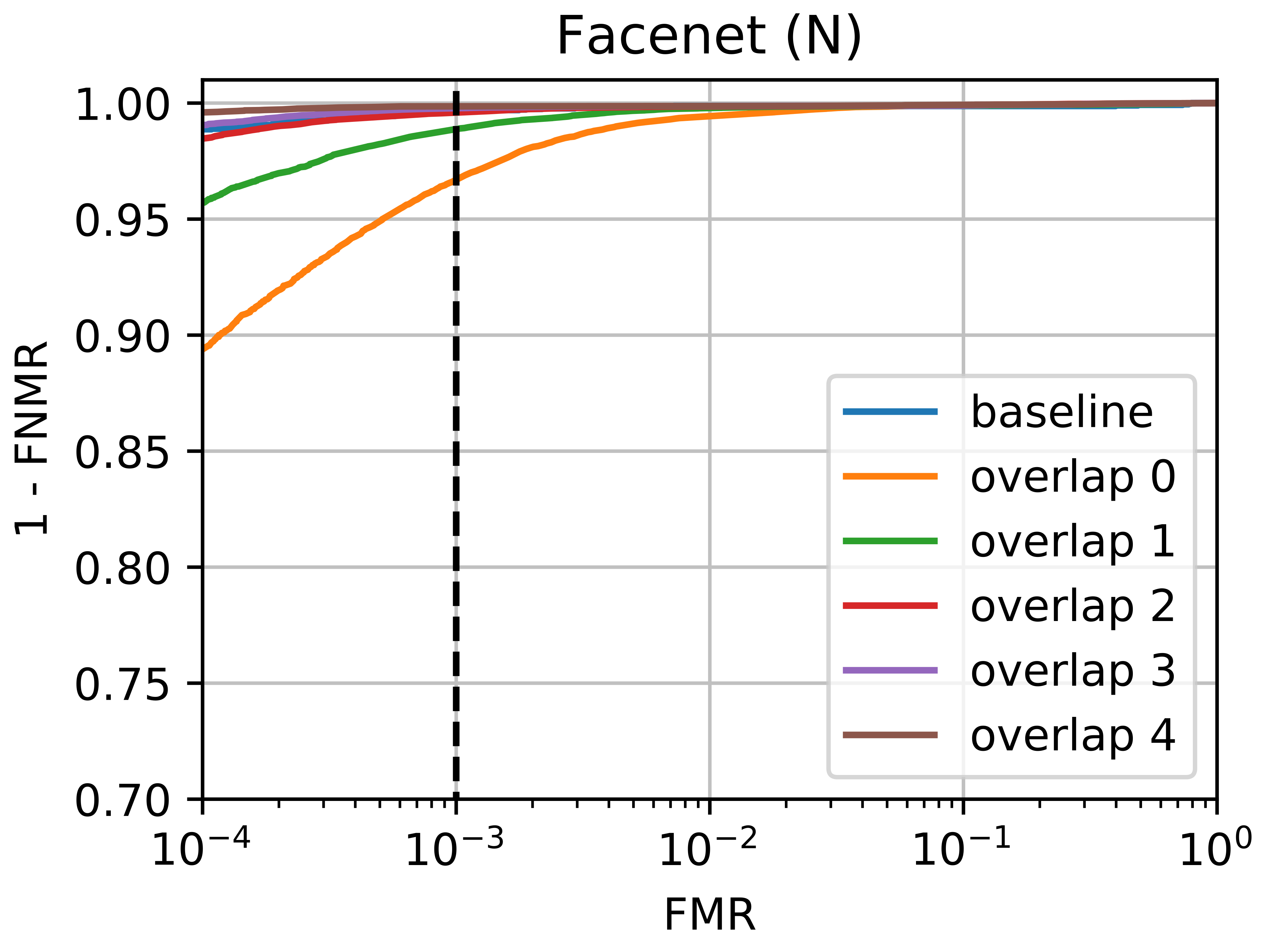}
\hfill
\includegraphics[width=0.49\columnwidth]{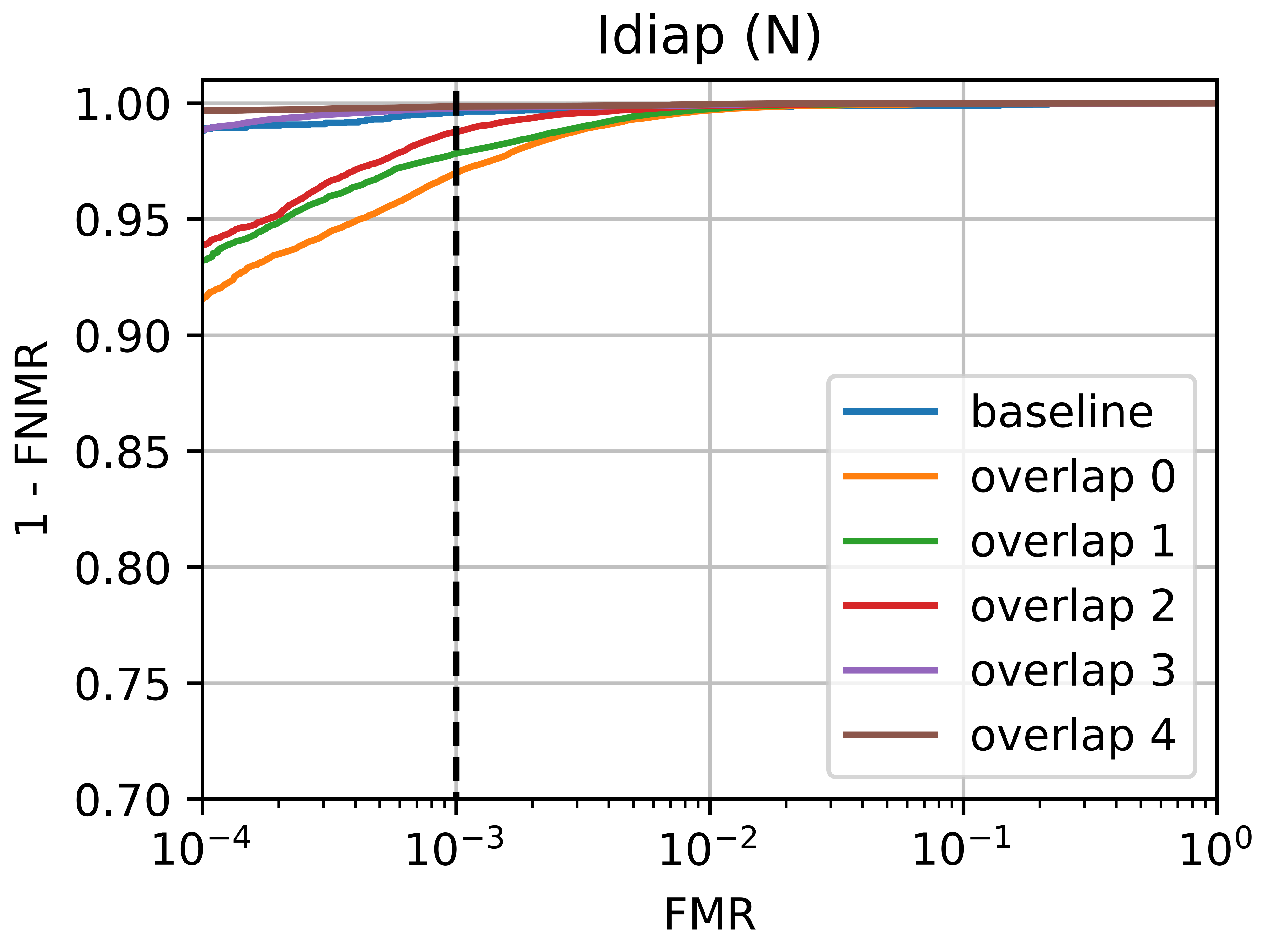}
\vfill
\includegraphics[width=0.49\columnwidth]{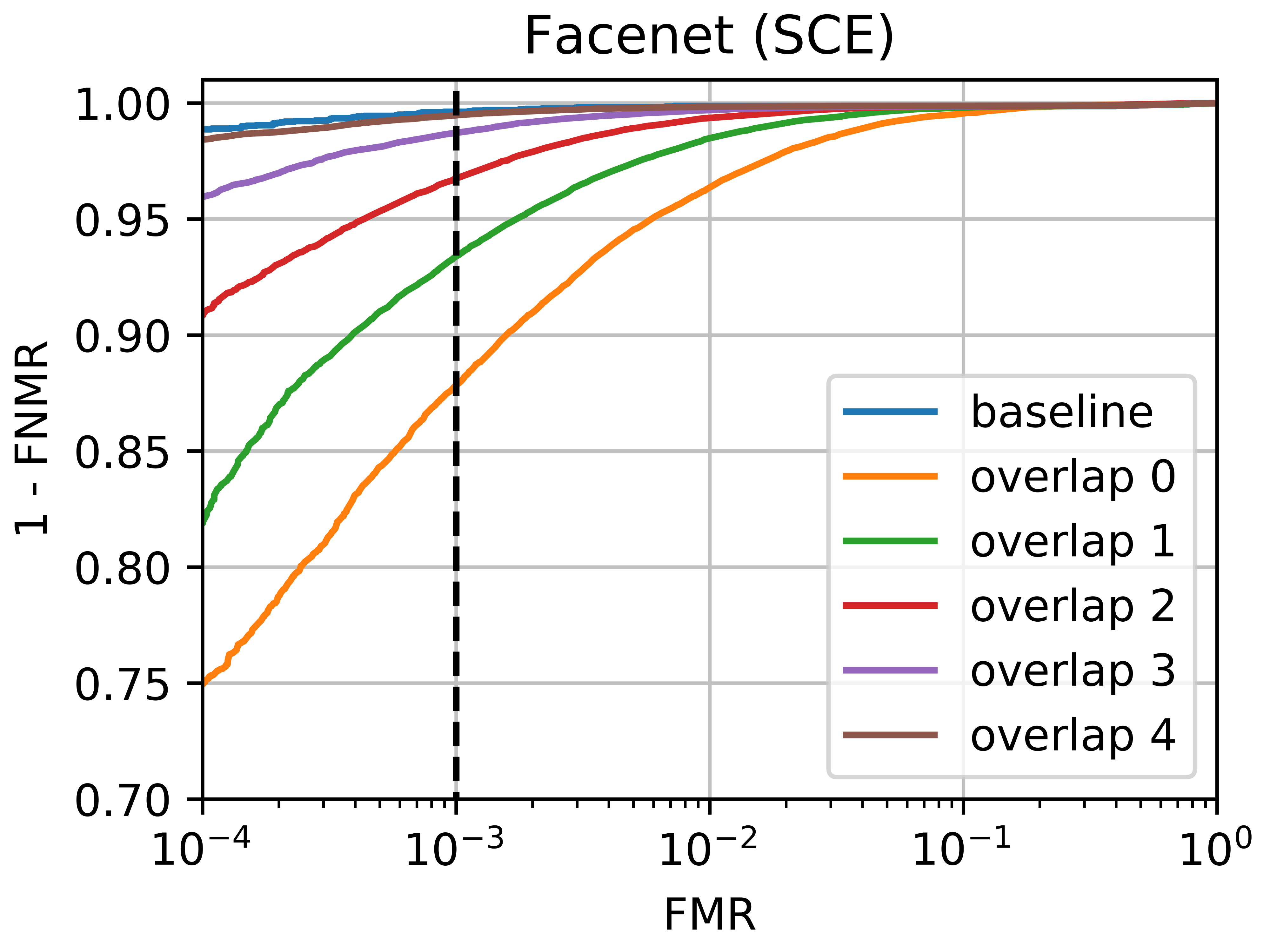}
\hfill
\includegraphics[width=0.49\columnwidth]{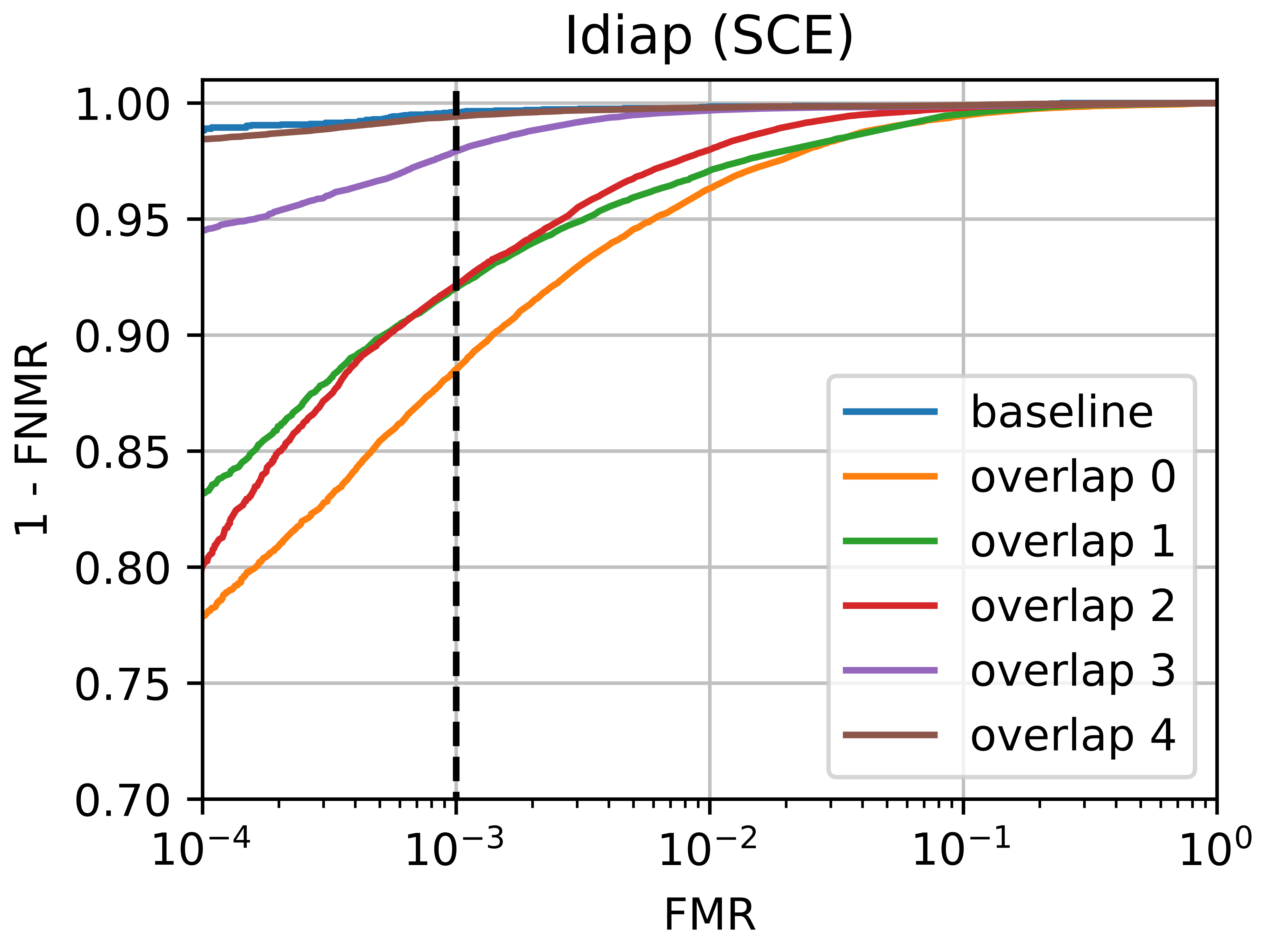}
\caption{ROC plots comparing the baseline and PolyProtected face verification systems, in the N and SCE scenarios. The vertical dashed line in each plot corresponds to the match threshold at a False Match Rate (FMR) of $10^{-3}$ or 0.1\%, which is a commonly used criterion.}
\end{figure}

Additionally, Table 1 summarises the True Match Rates (TMR = 1 - FNMR) from Fig. 4, at match thresholds corresponding to FMR = \{0.01\%, 0.1\%, 1\%\} (i.e., FMR = \{$10^{-4}$, $10^{-3}$, $10^{-2}$\}). We present only the TMRs for PolyProtect's N scenario, since this is its intended operational scenario.

\begin{table}[!h]
\renewcommand{\arraystretch}{1.3}
\caption{TMR at match thresholds corresponding to different FMR values, for the baseline and PolyProtected (N scenario) systems.}
\centering
\begin{tabular}{|c|c|c|c|c|c|}
\hline
\multicolumn{3}{|c}{\multirow{2}{*}{\textbf{System}}} & \multicolumn{3}{|c|}{\textbf{TMR @ FMR = }} \\
\cline{4-6}
\multicolumn{3}{|c|}{} & \textbf{0.01\%} & \textbf{0.1\%} & \textbf{1\%} \\
\hline
\multirow{6}{*}{Facenet} & \multicolumn{2}{c|}{\textit{Baseline}} & \textit{98.87\%} & \textit{99.62\%} & \textit{99.87\%} \\
\cline{2-6}
 & \multirow{5}{*}{\shortstack{PolyProtected:\\Overlap =}} & 0 & 89.39\% & 96.70\% & 99.45\% \\
 \cline{3-6}
 & & 1 & 95.69\% & 98.88\% & 99.79\% \\
 \cline{3-6}
 & & 2 & 98.47\% & 99.59\% & 99.85\% \\
 \cline{3-6}
 & & 3 & 99.05\% & 99.78\% & 99.87\% \\
 \cline{3-6}
 & & 4 & 99.59\% & 99.87\% & 99.87\% \\
\hline
\multirow{6}{*}{Idiap} & \multicolumn{2}{c|}{\textit{Baseline}} & \textit{98.80\%} & \textit{99.60\%} & \textit{99.85\%} \\
\cline{2-6}
 & \multirow{5}{*}{\shortstack{PolyProtected:\\Overlap =}} & 0 & 91.54\% & 96.98\% & 99.69\% \\
 \cline{3-6}
 & & 1 & 93.15\% & 97.82\% & 99.74\% \\
 \cline{3-6}
 & & 2 & 93.86\% & 98.76\% & 99.88\% \\
 \cline{3-6}
 & & 3 & 98.86\% & 99.78\% & 99.93\% \\
 \cline{3-6}
 & & 4 & 99.66\% & 99.85\% & 99.95\% \\
\hline
\end{tabular}
\end{table}

There are several important observations from Fig. 4 and Table 1. Firstly, in both the N and SCE scenarios, the recognition accuracy of the PolyProtected systems generally improves as the amount of overlap increases. This makes sense, because using a larger overlap in the PolyProtect mapping results in the generation of a higher-dimensional PolyProtected template (see Fig. 2), which contains more information from the original face embedding. So, the larger the overlap, the better the expected recognition accuracy.

Secondly, for the N scenario, at the most commonly used match threshold at FMR = $10^{-3}$ (or 0.1\%), marked by the vertical dashed line in Fig. 4, the recognition accuracy of both the Facenet and Idiap PolyProtected systems is slightly \textit{better} than that of the corresponding baseline, when an overlap of 3 or 4 is used in the PolyProtect mapping (see Table 1 for the exact values). This is despite the fact that these PolyProtected systems work with templates of lower dimensionality (63 and 124 for overlaps of 3 and 4, respectively) than that of the face embeddings used in the baseline systems (128). So, this improvement in the recognition accuracy of the aforementioned PolyProtected systems is most probably due to the use of user-specific $C$ and $E$ parameters in the generation of the PolyProtected templates, which increases the separation between different users in the protected feature domain. Although the recognition accuracy of the PolyProtected systems using overlaps of 0-2 is a little worse than that of the baseline for both Facenet and Idiap at the same match threshold, even the worst recognition accuracy (i.e., at overlap = 0) is still very good, resulting in a TMR of almost 97\%. Furthermore, if a higher FMR can be tolerated, then the difference in the recognition accuracy of all PolyProtected systems and the corresponding baseline becomes fairly insignificant, reaching approximate equivalence at an FMR of $10^{-2}$ (1\%). Finally, both Fig. 4 and Table 1 suggest that setting the match threshold to the stricter value corresponding to an FMR of $10^{-4}$ (0.01\%) results in a fairly significant degradation in the recognition accuracy of PolyProtected systems using an overlap of 0 (compared to the corresponding baseline). Although this drop in the recognition accuracy may be considered acceptable in some application scenarios, in practice the overlap parameter should be \textit{tuned} according to the desired ``recognition accuracy versus irreversibility'' trade-off (explored in the irreversibility analysis in Section 4.3). Having said that, we should emphasize that the aim of this paper is to evaluate PolyProtect in a cooperative-user mobile face verification context, which is the most likely operational scenario for PolyProtect in practice. In this case, a match threshold at an FMR of 0.01\% (or lower) is unlikely to be necessary, since we would not expect anywhere near 10,000 impostor verification attempts. In fact, a match threshold at an FMR of 0.1\% (corresponding to an impostor accept rate of 1 in 1,000) should be more than adequate for this application scenario, and applications where impostor access attempts are even less likely may adopt a more lenient match threshold (e.g., at an FMR of 1\%, which corresponds to 1 in 100 impostors being accepted).  

Thirdly, Fig. 4 indicates that, in the SCE scenario, the recognition accuracy of the PolyProtected systems is worse, in general, than the baselines. This may be attributed to the fact that, in this scenario, we are essentially performing a dimensionality reduction in the mapping from each subject's face embedding to their PolyProtected template, without the benefits of the additional user-specific information as in the N scenario. Consequently, the amount of discriminative information in the protected templates may be expected to be less than that in the unprotected face embeddings, resulting in lower recognition accuracy. Nevertheless, even in this worst-case scenario, the recognition accuracy of both the Facenet and Idiap PolyProtected systems using an overlap of 4 is almost equivalent to that of the corresponding baselines at FMR = $10^{-3}$. Even the PolyProtected systems using an overlap of 3 appear to perform very well, with a TMR (1 - FNMR) of around 0.98 (98\%) for Facenet and 0.97 (97\%) for Idiap. Furthermore, similarly to the observation made for the N scenario, a higher FMR tolerance in the employed systems would ensure that all PolyProtected systems (i.e., for all overlaps) achieve high recognition accuracy even in the SCE scenario; for example, at an FMR of $10^{-2}$ (1\%) in Fig. 4, the TMR for all PolyProtected systems is over 0.95 (95\%), and the performance of PolyProtected systems using overlaps of 3 and 4 achieves equivalence with the baselines. 

In summary, our analysis indicates that the recognition accuracy of a PolyProtected system depends on the amount of overlap used in the PolyProtect mapping, with the accuracy improving as the overlap increases. When the system operates in the envisioned (N) scenario, the recognition accuracy may be expected to be higher than, approximately equivalent to, or not significantly worse than, that of the baseline system (using unprotected face embeddings), depending on the chosen match threshold. In the worst-case (SCE) scenario where all users' parameters are stolen, the recognition accuracy of a PolyProtected system may be worse than that of the baseline; however, our results indicate that the performance can be improved by tuning the amount of overlap used in the PolyProtect mapping as well as the FMR tolerance of the underlying system. So, although this worst-case scenario is highly unlikely to occur in practice\footnote{It is unlikely that each user would steal \textit{all} the other users' PolyProtect parameters, especially at the same time. So, the N scenario results are more indicative of the expected recognition accuracy.}, even in this case the recognition accuracy should be acceptable, ensuring that the system does not suffer much in the time it takes to replace the compromised PolyProtected template(s). We may thus reasonably conclude that PolyProtect is capable of satisfying the \textit{recognition accuracy} property of a face embedding protection scheme.

It would be useful to present quantified comparisons of the recognition accuracy of PolyProtect to that of other face embedding protection schemes in the literature, but it is currently not possible to do this fairly. For the comparison to be as fair as possible, the following conditions should be satisfied: (1) the methods should be evaluated on the \textit{same dataset}, and to ensure that the comparison remains relevant to the context of the PolyProtect evaluation presented in this paper, this dataset should be Mobio; (2) the raw embeddings should be generated using the \textit{same feature extraction} process, and we should be able to \textit{de-couple} the feature extraction and protection steps, to ensure that the comparison targets \textit{solely} the protection algorithms; (3) we should have access to the methods' implementation code, to allow for better insight into whether or not a fair comparison would be possible. Condition (1) is not satisfied, because none of the existing face embedding protection methods have been evaluated on the Mobio dataset. Since the aim of this paper is to evaluate PolyProtect's suitability for adoption in a mobile face verification scenario, and as no other face embedding protection method has been evaluated on a database that is equally/more suitable (compared to Mobio) for representing this target scenario, it is not possible to present a fair numerical comparison against the recognition accuracy results reported for the existing protection methods. Conditions (2) and (3) are satisfied only by \cite{b18}, which employs fully homomorphic encryption to protect 128-dimensional embeddings generated using the same open-source FaceNet model adopted for our Facenet baseline. A known property of homomorphic encryption methods is their ability to ensure approximately zero loss in the recognition accuracy; indeed, \cite{b18} showed that this is achievable regardless of the evaluation dataset, depending on the precision of the face embedding quantisation scheme and the adopted match threshold. So, without an explicit comparison on the Mobio dataset, we may reasonably conclude that PolyProtect appears comparable to \cite{b18} in its ability to be tuned to approximately maintain the recognition accuracy of the baseline (unprotected) system, when the protected system operates as intended (N scenario for PolyProtect, and the use of user-specific encryption/decryption keys for \cite{b18}).
 
Having made this comparison, we also note that the use of PolyProtect may actually cause an \textit{increase} in the resulting recognition accuracy (as discussed in our analysis of Fig. 4 and Table 1), while this would not be possible using the homomorphic encryption method in \cite{b18}. On the other hand, although \cite{b18} does not present the equivalent of PolyProtect's SCE scenario (i.e., stolen encryption/decryption keys), we may expect that the use of \textit{same} encryption/decryption keys to enroll \textit{different} subjects in \cite{b18} would produce approximately the same comparison scores as for the \textit{user-specific} key scenario. Despite this potential advantage of \cite{b18} over PolyProtect in the worst-case (albeit practically unlikely) scenario, the main pitfall of \cite{b18} is that the face embeddings are secure only insofar as the decryption keys remain secret. PolyProtected templates, on the other hand, do not rely solely on the secrecy of the user-specific parameters to protect the original face embeddings. An analysis of the irreversibility of PolyProtect is presented in Section 4.3.

\subsection{Irreversibility}

A face embedding protection method is considered irreversible (non-invertible) if it is impossible (or computationally infeasible) to recover the original (unprotected) face embedding from its protected template. This section investigates the irreversibility of PolyProtect. Section 4.3.1 defines our adopted threat model, on which the subsequent irreversibility analysis will be based. We then consider the difficulty of recovering the original face embedding from one (Section 4.3.2) or more (Section 4.3.3) PolyProtected templates from the same subject.

\subsubsection{Threat model}

To analyse PolyProtect's irreversibility, we must first define our \textit{threat model}, as specified in ISO/IEC 30316 (the international standard on performance testing of biometric template protection schemes)\footnote{\url{https://bit.ly/3hLRnYM}}. The threat model characterises the type of attacker on which we wish to base our irreversibility analysis. The most difficult threat model in ISO/IEC 30316 is referred to as a \textit{full disclosure model}, which assumes that the attacker knows everything there is to know about the protection method (e.g., algorithms, secrets, etc.). Since this type of attacker would represent the worst-case scenario in practice, we decided to base our analysis of the irreversibility of PolyProtect on the \textit{full disclosure model}.

In the context of PolyProtect, we define the full disclosure threat model as assuming that the attacker has access to the following information: knowledge of the PolyProtect algorithm, including the number of embedding elements ($m$) used to generate each PolyProtected element, the amount of \textit{overlap} used in the PolyProtect mapping, as well as the user-specific $C$ and $E$ parameters that define PolyProtect's polynomials; one or more PolyProtected templates, $P$, corresponding to a particular face embedding, $V$; and knowledge of a face embedding element distribution, which is representative of the face embeddings used to create the PolyProtected templates to which the attacker has access. The attacker's goal, therefore, is to use all this information to attempt to recover a subject's original face embedding, $V$, from one or more of their PolyProtected templates, $P$.

Recall that, in our experimental set-up (see Section 4.1), a face embedding consists of 128 elements, i.e., $V = [v_1, v_2, ..., v_{128}]$. In the PolyProtect irreversibility analysis, the attacker's aim is to recover these 128 elements from the corresponding PolyProtected template, $P = [p_1, p_2, ..., p_k]$ (where $k$ depends on the amount of overlap used in the PolyProtect mapping, as illustrated in Fig. 2). Since our full disclosure threat model assumes that the attacker has some knowledge of the distribution of these 128 embedding elements, the first step in analysing the irreversibility of PolyProtect was to estimate this distribution. Note that, as defined by the Mobio database protocol we adopted, the face embeddings used in our experiments were divided into two sets: \textit{development} and \textit{evaluation}. Each set was further split into \textit{reference} embeddings (used for enrollment) and \textit{query} embeddings (used for verification). Since the intention of the adopted database protocol is to report results on the evaluation set, we may consider the evaluation set's reference face embeddings as the embeddings that are used to generate the \textit{enrolled} PolyProtected templates. It is, therefore, precisely these embeddings that the attacker in our irreversibility analysis is trying to recover. This means that the attacker should \textit{not} have access to these embeddings; however, they may be assumed to have access to the development set, since these embeddings are not used for enrollment\footnote{In general, the purpose of the development set is to establish certain evaluation parameters, such as the match threshold.}. The attacker could, therefore, use the \textit{development} set's reference face embeddings to estimate the distribution of the 128 elements in the \textit{evaluation} set's reference embeddings, which they are attempting to recover.

Our irreversibility analysis thus began with an estimation of the probability distribution for each of the 128 elements in a face embedding, using the reference embeddings from Mobio's \textit{development} dataset, separately for our Facenet and Idiap baseline systems. This means that each embedding element was considered separately across all the reference embeddings, to estimate its corresponding probability distribution. So, for each of our two baseline systems, we ended up with 128 different probability distributions, one for each of the 128 embedding elements. 

Note that these probability distributions reveal a lot of information about the underlying embeddings (and thus the face images from the Mobio database), so we are unable to publish the distributions in this paper, because access to the Mobio dataset is granted based on an end-user license agreement. Since our PolyProtect code is publicly available, the interested reader may generate these results for themselves upon gaining lawful access to Mobio.    

The probability distributions were next used to estimate the irreversibility of PolyProtect when the fully-informed attacker has access to one (Section 4.3.2) or more (Section 4.3.3) PolyProtected templates from the same person. 

\subsubsection{Access to single PolyProtected template}

This section considers the feasibility of recovering the original face embedding, $V$, from its PolyProtected template, $P$. A fully-informed attacker could attempt this $P \rightarrow V$ inversion in one of two ways: (i) analytically, or (ii) numerically.

An analytical approach towards inverting $P$ would involve re-arranging the system of multivariate polynomial equations used in the $V \rightarrow P$ mapping (see Section 3), to obtain an explicit solution for each variable, $v_i$ ($i = 1, 2, ..., n$), in $V = [v_1, v_2, ..., v_n]$. Note that a \textit{variable} corresponds to an element in the original face embedding, so in the case of our experimental set-up, which employed 128-dimensional face embeddings, there are 128 variables, i.e., $V = [v_1, v_2, ..., v_{128}]$. This means that our $V \rightarrow P$ mapping is a mapping from a 128-dimensional space to a lower-dimensional space, where the dimensionality of the PolyProtected template, $k$, is determined by the amount of overlap used (see Fig. 2). Consequently, the inverse mapping, $P \rightarrow V$, would be a mapping from the $k$-dimensional space to the 128-dimensional space. These forward and inverse mappings are summarised in Table 2.

\begin{table}[!h]
\renewcommand{\arraystretch}{1.3}
\caption{$V \rightarrow P$ and $P \rightarrow V$ mappings for different overlap amounts.}
\centering
\begin{tabular}{|c|c|c|}
\hline
\textbf{Overlap} & $\mathbf{V \rightarrow P}$ & $\mathbf{P \rightarrow V}$ \\
\hline
0 & $\mathbb{R}^{128} \rightarrow \mathbb{R}^{26}$ & $\mathbb{R}^{26} \rightarrow \mathbb{R}^{128}$ \\
\hline 
1 & $\mathbb{R}^{128} \rightarrow \mathbb{R}^{32}$ & $\mathbb{R}^{32} \rightarrow \mathbb{R}^{128}$ \\
\hline
2 & $\mathbb{R}^{128} \rightarrow \mathbb{R}^{42}$ & $\mathbb{R}^{42} \rightarrow \mathbb{R}^{128}$ \\
\hline
3 & $\mathbb{R}^{128} \rightarrow \mathbb{R}^{63}$ & $\mathbb{R}^{63} \rightarrow \mathbb{R}^{128}$ \\
\hline
4 & $\mathbb{R}^{128} \rightarrow \mathbb{R}^{124}$ & $\mathbb{R}^{124} \rightarrow \mathbb{R}^{128}$ \\
\hline
\end{tabular}
\end{table}

Considering Table 2 alongside the description of PolyProtect in Section 3, we can interpret the presented mappings as follows. When an overlap of 0 is used, the $V \rightarrow P$ mapping is defined by 26 equations in 128 variables (unknowns), resulting in a 26-dimensional PolyProtected template, $P$. Based on the description of PolyProtect in Section 3, we know that each of these 26 equations consists of a unique set of 5 variables. So, it is reasonable to assume that the 26 equations are linearly independent, meaning that the inverse mapping ($P \rightarrow V$) cannot be uniquely defined, due to the $128 - 26 = 102$ degrees of freedom.  

Similar observations can be made for overlaps 1 to 4; however, in this case, each equation does not consist of an entirely unique set of variables, since there is \textit{some} overlap between the different sets of embedding elements used to generate each PolyProtected element, meaning that there are fewer degrees of freedom in the inverse mapping. Nevertheless, since every equation consists of at least one variable that is not used by any of the other equations, it is still reasonable to assume that all the equations, for all overlaps, are linearly independent. Consequently, we may conclude that the inverse mapping, $P \rightarrow V$, is defined by an \textit{underdetermined} system of equations and, therefore, technically does not exist. This is because there are (theoretically) infinitely many solutions for the elements in $V$ that could produce $P$, so there is \textit{no unique solution}.

Furthermore, as discussed in Section 4.1, the Abel-Ruffini theorem states that there is no closed-form algebraic expression for solving polynomials of degree 5 or higher. Since we used $m = 5$ in our PolyProtect implementation, each equation in the $V \rightarrow P$ mapping is a multivariable 5-degree polynomial. So, we may deduce that an analytical solution for $P \rightarrow V$ does \textit{not} exist, implying that PolyProtected templates are \textit{theoretically irreversible}.

Although PolyProtect is irreversible in theory, in practice there will not be an infinite number of solutions for $V$. This is because the values of $V$ will be limited in some way (e.g., by the range and precision of possible values) depending on the implementation of the underlying face recognition system. Consequently, the number of valid solutions will be constrained in practice. Furthermore, the mathematical impossibility of deriving an \textit{exact} solution to the problem of recovering $V$ from $P$ may present a smaller obstacle in practice, where an attacker might be satisfied with an \textit{approximation} of $V$. So, although the inverse mapping, $P \rightarrow V$, cannot be explicitly defined, a real-world attacker may attempt to use a numerical solver to converge to an approximate solution for $V$ from a set of initial guesses. 

To estimate the feasibility of such an attempt in practice, we simulated this irreversibility attack using an open-source numerical solver: Python's \textit{scipy.optimize.root} function with the \textit{lm} method. This method adopts the Levenberg-Marquardt algorithm, which approximates a solution to a non-linear system of equations using a damped least-squares approach. This algorithm is reported to be very reliable for solving non-linear, medium-sized (i.e., a few hundred variables) optimization problems in practice\footnote{\url{https://bit.ly/3tSGKbo}}, and can be applied to underdetermined systems of equations\footnote{\url{https://bit.ly/3nQcvkc}}. Since the recovery of $V$ from $P$ is represented as an underdetermined 128-variable system of non-linear equations, we deemed this method very suitable for our purposes.    

The first step in attempting to recover an approximation of the original face embedding, $V$, from its PolyProtected template, $P$, using the aforementioned numerical solver, was to set up a system of $k$ equations (where $k$ corresponds to the dimensionality of $P$, as per Table 2) for each overlap amount.  For example, the system of 26 equations for overlap = 0 (see Fig. 2) was set up for the numerical solver as follows:

\begin{align*}
	c_{1}v_{1}^{e_1} + c_{2}v_{2}^{e_2} + c_{3}v_{3}^{e_3} + c_{4}v_{4}^{e_4} + c_{5}v_{5}^{e_5} - p_{1} &= 0 \\
	\\
	c_{1}v_{6}^{e_1} + c_{2}v_{7}^{e_2} + c_{3}v_{8}^{e_3} + c_{4}v_{9}^{e_4} + c_{5}v_{10}^{e_5} - p_{2} &= 0 \\
	& \vdots \\
	c_{1}v_{126}^{e_1} + c_{2}v_{127}^{e_2} + c_{3}v_{128}^{e_3} + c_{4}0^{e_4} + c_{5}0^{e_5} - p_{26} &= 0
\end{align*}

Note the use of zeros in place of $v_{129}$ and $v_{130}$ in the last (26\textsuperscript{th}) equation, which is due to the padding required to make the 128 dimensions of our face embeddings divisible by 5 (our choice for $m$, as explained in Section 4.1). The systems of equations for overlaps 1 to 4 were set up in a similar way, except that each system consisted of a different number of equations (as per Table 2). For example, for overlap = 4, the system consisted of 124 equations, as follows:

\begin{align*}
	c_{1}v_{1}^{e_1} + c_{2}v_{2}^{e_2} + c_{3}v_{3}^{e_3} + c_{4}v_{4}^{e_4} + c_{5}v_{5}^{e_5} - p_{1} &= 0 \\
	\\
	c_{1}v_{2}^{e_1} + c_{2}v_{3}^{e_2} + c_{3}v_{4}^{e_3} + c_{4}v_{5}^{e_4} + c_{5}v_{6}^{e_5} - p_{2} &= 0 \\
	& \vdots \\
	c_{1}v_{124}^{e_1} + c_{2}v_{125}^{e_2} + c_{3}v_{126}^{e_3} + c_{4}v_{127}^{e_4} + c_{5}v_{128}^{e_5} - p_{124} &= 0
\end{align*}

Since it was possible to fully formulate each of the 124 equations using the 128 embedding elements ($v_{1}, v_{2}, ..., v_{128}$), no padding of the face embeddings was necessary in this case. Please refer to Fig. 2 to visualise the systems of equations for all overlap amounts.

Next, the numerical solver was used to approximate a solution for $V = [v_1, v_2, ..., v_{128}]$ from every $P$ that was generated for the \textit{evaluation} set of reference face embeddings in Section 4.2 (i.e., a total of 380 $P$ templates for each overlap amount). Since the solver requires a set of initial guesses for $v_1, v_2, ..., v_{128}$, we randomly generated 100 guesses for each of the 128 embedding elements. The guesses were drawn from the probability distributions established for the corresponding embedding elements on the \textit{development} set of reference embeddings (as explained in Section 4.3.1). 

The numerical solver then started from these initial guesses to estimate a solution for each of the 128 elements in $V$, from the corresponding $P$. As soon as the solver indicated that a solution for $V$ had been found (by setting the Boolean flag \texttt{success} to 1), the process was stopped. Since the error tolerance for the solution was set to a very small (default) value ($1.49012e-08$), an attacker attempting this inversion attack in practice should have no reason to doubt that the solution returned at this point in the process is a good approximation to $V$. So, not all 100 initial guesses necessarily needed to be tried. Once all 380 $P$ templates had passed through the solver, we calculated the \textit{solution rate} as the proportion of all $P$s for which a solution was found.

If a solution, $V^{*}$, for a particular $P$ was found, we then calculated the comparison score between this approximation of the corresponding face embedding and the true embedding, $V$, in terms of the negative cosine distance (as for the recognition accuracy in Section 4.2). The idea was to determine whether $V^{*}$ was a \textit{close enough} approximation to $V$, such that $V^{*}$ could be used to launch a replay attack in our baseline face recognition systems, which store the \textit{unprotected} $V$ as a reference face embedding. Since the closeness of a match between two face embeddings in practice would always be dependent on a match threshold (i.e., the match will never be perfect), it makes sense to judge the success of an inversion attack in the same way. 

We used the aforementioned approach to calculate the \textit{match rate} for each set of $P$ templates for which a solution for $V$ was found by the numerical solver. The match rate refers to the proportion of $P$s for which the comparison score between $V^{*}$ and $V$ was greater than or equal to a pre-defined match threshold. The match rate was computed at three thresholds established on the baseline systems' \textit{development} set\footnote{Simulates the real-life scenario where match thresholds are tuned offline, on a different set of subjects to that encountered in the systems' deployment scenario (represented by the \textit{evaluation} set of embeddings).} of face embeddings: at FMR = 0.01\% (fairly strict), at FMR = 0.1\% (commonly used), and at FMR = 1\% (fairly lenient), to represent the progression from a higher-security to a lower-security application scenario.

Finally, we calculated the \textit{inversion success rate} = \textit{solution rate} $\times$ \textit{match rate}, to estimate the overall success rate for an attacker attempting the $P \rightarrow V$ inversion using the numerical solver. Table 3 presents the resulting inversion success rates for our Facenet and Idiap PolyProtected systems.

\begin{table}[!h]
\renewcommand{\arraystretch}{1.3}
\caption{Inversion success rates at different match thresholds.}
\centering
\begin{tabular}{|c|c||c|c|}
\hline
\textbf{Threshold} & \textbf{Overlap} & \textbf{Facenet} & \textbf{Idiap} \\
\hline
\multirow{5}{*}{@ FMR = 0.01\%} & 0 & 0.00 & 0.00 \\

    & 1 & 0.00 & 0.00 \\
    
    & 2 & 0.00 & 0.00 \\
    
    & 3 & 0.05 & 0.02 \\
    
    & 4 & 0.95 & 0.96 \\
\hline
\multirow{5}{*}{@ FMR = 0.1\%} & 0 & 0.00 & 0.00 \\

    & 1 & 0.00 & 0.00 \\

    & 2 & 0.01 & 0.01 \\

    & 3 & 0.15 & 0.15 \\

    & 4 & 0.95 & 0.96 \\
\hline
\multirow{5}{*}{@ FMR = 1\%} & 0 & 0.00 & 0.00 \\

    & 1 & 0.01 & 0.01 \\
    
    & 2 & 0.08 & 0.05 \\
    
    & 3 & 0.52 & 0.49 \\
    
    & 4 & 0.95 & 0.97 \\
\hline
\end{tabular}
\end{table}

Table 3 shows two important trends. Firstly, the inversion success rate is, in general, \textit{lower} when the baseline systems operate at a \textit{stricter} match threshold (at a lower FMR). This is because a stricter threshold would impose a tougher standard on what is meant by a \textit{good enough} approximation of $V$. So, the stricter the threshold in practice, the less likely the inversion attack would be to succeed, in the sense that it would become more difficult to match the recovered approximation of $V$ (i.e., $V^{*}$) to the unprotected $V$ used in the baseline face recognition system. This trend is most evident for an overlap of 3, for which the inversion success rate for both the Facenet and Idiap systems is close to 0 at the strictest match threshold (at FMR = 0.01\%), equal to 0.15 at the commonly used threshold (at FMR = 0.1\%), and around 0.5 at the most lenient threshold (at FMR = 1\%).

The second important trend from Table 3 is that the inversion success rate tends to \textit{increase} as the amount of overlap used in the PolyProtect mapping \textit{increases}. This makes sense, because a larger overlap results in a higher-dimensional PolyProtected template (see Table 2 and Fig. 2), which contains a greater amount of information from the original face embedding. Consequently, the system of equations in the $P \rightarrow V$ mapping for larger overlaps is more constrained than for smaller overlaps, meaning that, if a solution for $V$ is found, it is more likely to be closer to the true solution. For example, we see that for both the Facenet and Idiap systems, the inversion success rate for PolyProtected templates generated using overlaps of 0-2 is close to 0 at all three match thresholds, meaning that the templates can be considered \textit{practically irreversible} at these operating points. Although the inversion success rate for an overlap of 3 was found to be higher, the results indicate that the attacker would still \textit{fail} to invert the PolyProtected template 95\% and 98\% of the time for the Facenet and Idiap systems, respectively, when the baseline match threshold is set at FMR = 0.01\%, 85\% of the time for both systems when the threshold is set at FMR = 0.1\%, and about 50\% of the time when the threshold is set at FMR = 1\%. This is still significantly better than using the original (unprotected) face embeddings, especially considering that this evaluation was based on the assumption of a \textit{fully-informed attacker}, which should be extremely rare in practice. 

Overall, our analysis suggests that using overlaps of 0, 1, or 2 would be the safest option, and that an overlap of 3 might be acceptable depending on the unprotected system's operating match threshold. An overlap of 4, on the other hand, appears to be an unwise choice, because the inversion success rate was found to be close to 1 for all three thresholds in Table 3. Having said this, recall that our analysis is based on the toughest threat model, whereby we assume that our simulated attacker knows \textit{everything} there is to know about the PolyProtected systems, including representative probability distributions of the original face embedding elements, as well as all the parameters employed in the PolyProtect mapping. This should \textit{not} be the case in practice, meaning that a real-life inversion attack should be significantly less likely to succeed. For example, in practice the user-specific parameters, $C$ and $E$, should \textit{not} be stored in the open, but rather secured separately (e.g., via encryption, on a separate token, etc.). Also, it should be difficult for the attacker to obtain representative embedding distributions from which to draw the initial guesses for the numerical solver. These factors combined would make it much more difficult for the solver to converge to a close approximation of the original embedding, meaning that, in a real-life inversion attack, we would expect the inversion success rate to be close to 0 most of the time.

Considering once again our worst-case evaluation scenario based on a fully-informed attacker, the two main trends observed in Table 3 imply that the inversion success rate of PolyProtected templates in practice will depend on the amount of overlap used in the PolyProtect mapping and the operating threshold of the baseline face recognition system in which the unprotected face embeddings are employed. If we were to form a general recommendation on what overlap amount to use for the PolyProtect mapping in practice (for the considered evaluation scenario), we should consider Table 3 alongside the trends shown by the ROC plots in Fig. 4 (and with the help of Table 1). Based on this comparison, we may conclude that, in general, an overlap of 2 would most likely provide the best trade-off between irreversibility and recognition accuracy. Since the corresponding PolyProtected templates were shown to be practically irreversible even for the worst-case (albeit unlikely) scenario of a fully-informed attacker, we may reasonably conclude that PolyProtect would be suitable for securing face embeddings in practice. Table 3 and Fig. 4 (with Table 1) further suggest that overlaps of 0 and 1 may be more suitable for lower-security PolyProtected applications, which operate at more lenient thresholds (i.e., corresponding to higher FMRs), while an overlap of 3 could offer the best accuracy versus irreversibility trade-off in higher-security PolyProtected applications (where a lower FMR is desired). Remember, however, that the \textit{smaller} the overlap, the more \textit{irreversible}, and thus more \textit{privacy-preserving}, the PolyProtected templates would be.

A meaningful, quantifiable comparison between the irreversibility of PolyProtect and that of the other face embedding protection schemes in the literature is currently extremely difficult. This is mainly due to inconsistencies in the adopted evaluation methodologies (there is no standardised approach, since the evaluation tends to be method-specific) and the assumed threat model (which is usually not even explicitly defined). Furthermore, it is often not evident how we should fairly select comparable parameters (e.g., user-specific transformation keys) across different protection methods, since such parameters (if they exist) do not always take the same form or have the same meaning. For such reasons, comparing the irreversibility of different face embedding protection methods in the literature may result in ambiguous conclusions on which method is better in this regard. So, irreversibility comparisons are usually avoided.    

On a final note, we would like to make a case for \textit{practical} irreversibility evaluation methodologies. Recall that the irreversibility analysis presented earlier for PolyProtect was based on defining a \textit{successful inversion} as the ability to use the approximation of $V$ recovered from its $P$ (i.e., $V^{*}$) to impersonate the same identity in an unprotected face recognition system that employs $V$ itself. This seems like a reasonable approach for estimating PolyProtect's irreversibility \textit{in practice}, because it gives us direct insight into what exactly a successful inversion attack might mean in a real-life scenario. In our view, this is more useful (from a practical point of view) than a theoretical evaluation, since such evaluations tend to be based on unrealistic assumptions (like the assumption of the existence of an infinite number of solutions for $V$, as discussed earlier in the context of an \textit{analytical} irreversibility analysis for PolyProtect). We thus hope to encourage a greater focus on practical evaluation methodologies when analysing the irreversibility of face embedding protection methods in the literature, particularly in terms of inversion attacks for which readily-available tools can be used (such as Python's numerical solver in our analysis). This would help to produce more tangible results, which may make it easier to understand and compare the expected irreversibility of different protection methods in practice. Furthermore, since we are at the stage where the protection of face embeddings is becoming an urgent requirement in real-life applications, a practically-oriented irreversibility analysis would surely be appreciated by deployers of face recognition systems. 

\subsubsection{Access to multiple PolyProtected templates}

Section 4.3.2 analysed the irreversibility of PolyProtect when an attacker has access to only one PolyProtected template, $P$, corresponding to a certain face embedding, $V$. In this section, we consider the scenario where the attacker has access to \textit{multiple} PolyProtected templates from the same $V$, which they attempt to combine to recover an approximation of $V$. This is referred to as a Record Multiplicity Attack (ARM) in the literature. This type of attack could occur in the scenario where the same face embedding is used to generate different PolyProtected templates (using different $C$ and $E$ parameters), then each PolyProtected template is either enrolled in a different application or used to replace a compromised PolyProtected template in the same application. Although such an attack should be extremely difficult to launch in practice\footnote{The attacker would need to know which applications to target and how to hack them to steal the subject's PolyProtected templates.}, it must still be considered when analysing PolyProtect's irreversibility.

To analyse the susceptibility of PolyProtect to ARM, we proceeded as follows. We assumed that different PolyProtected templates from the same subject are generated using the \textit{same} original face embedding, $V$. In practice, we should never use \textit{exactly the same} face embedding to generate different PolyProtected templates for the same subject. This is because the original face embedding should never be stored in the clear, meaning that each enrollment should request \textit{a new sample} of the enrollee's face. Consequently, each PolyProtected template belonging to the same subject should be generated using a different instance of $V$. In our ARM analysis, however, we consider the worst-case scenario where the \textit{same} $V$ is used to generate all of a subject's $P$ templates. This should make it easier to combine the information from these PolyProtected templates to recover the original face embedding, which means that this represents the best-case scenario for the attacker.  

The face embeddings and corresponding PolyProtected templates used in this analysis were the same as those used for the irreversibility analysis in Section 4.3.2. This means that each $V$ was associated with 10 different $P$s (each generated using different $C$ and $E$ parameters). Our ARM analysis thus consisted of attempting to recover an approximation of $V$ using 1 to 10 of its corresponding $P$s. This was simulated using the numerical solver approach explained in Section 4.3.2. This time, however, we were trying to solve systems of $k \times p$ equations (where $k$ is the dimensionality of the $P$ templates, and $p$ is the number of $P$s that the attacker is assumed to have access to), instead of only $k$ equations as in Section 4.3.2. 

So, the first step was to set up a system of $k \times p$ equations for each overlap amount\footnote{The $k$ values for overlaps 0 to 4 are indicated in Table 2.}, and for each value of $p$ in the range [1, 10]. For example, for overlap = 2, to simulate an attacker with access to the minimum of only \textit{one} 42-dimensional $P$, we set up a system of 42 equations in 128 unknowns (embedding elements), same as for the irreversibility analysis in Section 4.3.2. On the other hand, to simulate an attacker with access to the maximum of \textit{ten} 42-dimensional $P$s generated from the same $V$, we set up a system of 420 equations in 128 unknowns (i.e., one set of 42 equations used in the $V \rightarrow P$ mapping for each of the 10 PolyProtected templates). Note that, unlike the 42-equation system, the 420-equation system would not be \textit{underdetermined} but \textit{overdetermined}, meaning that the additional constraints to the solution space may allow the attacker to have a greater chance of recovering a good approximation of $V$ from ten $P$s than from one $P$. So, we would expect to see an increase in PolyProtect's susceptibility to ARM as the number of $P$s considered in the attack increases. However, an overdetermined system of equations may introduce inconsistencies to the solution space, so this expected trend cannot be considered certain in practice.

Similarly to the analysis in Section 4.3.2, we evaluated the \textit{inversion success rate} = \textit{solution rate} $\times$ \textit{match rate}. This was calculated separately for each $p$ in the range [1, 10] (i.e., 1 to 10 PolyProtected templates per $V$), and for each overlap amount. Note that, since the aim of this analysis was to determine whether using multiple PolyProtected templates would make it easier to recover the original face embedding than using only one PolyProtected template, the choice of match threshold was not so important, particularly because the irreversibility analysis in Section 4.3.2 already demonstrated the effects of using different match thresholds. So, Fig. 5 illustrates the inversion success rate for the Facenet and Idiap systems as the number of PolyProtected templates increases from 1 to 10, based only on the most commonly used match threshold at FMR = 0.1\% from Section 4.3.2. 

\begin{figure}[!h]
\centering
\includegraphics[width=0.49\columnwidth]{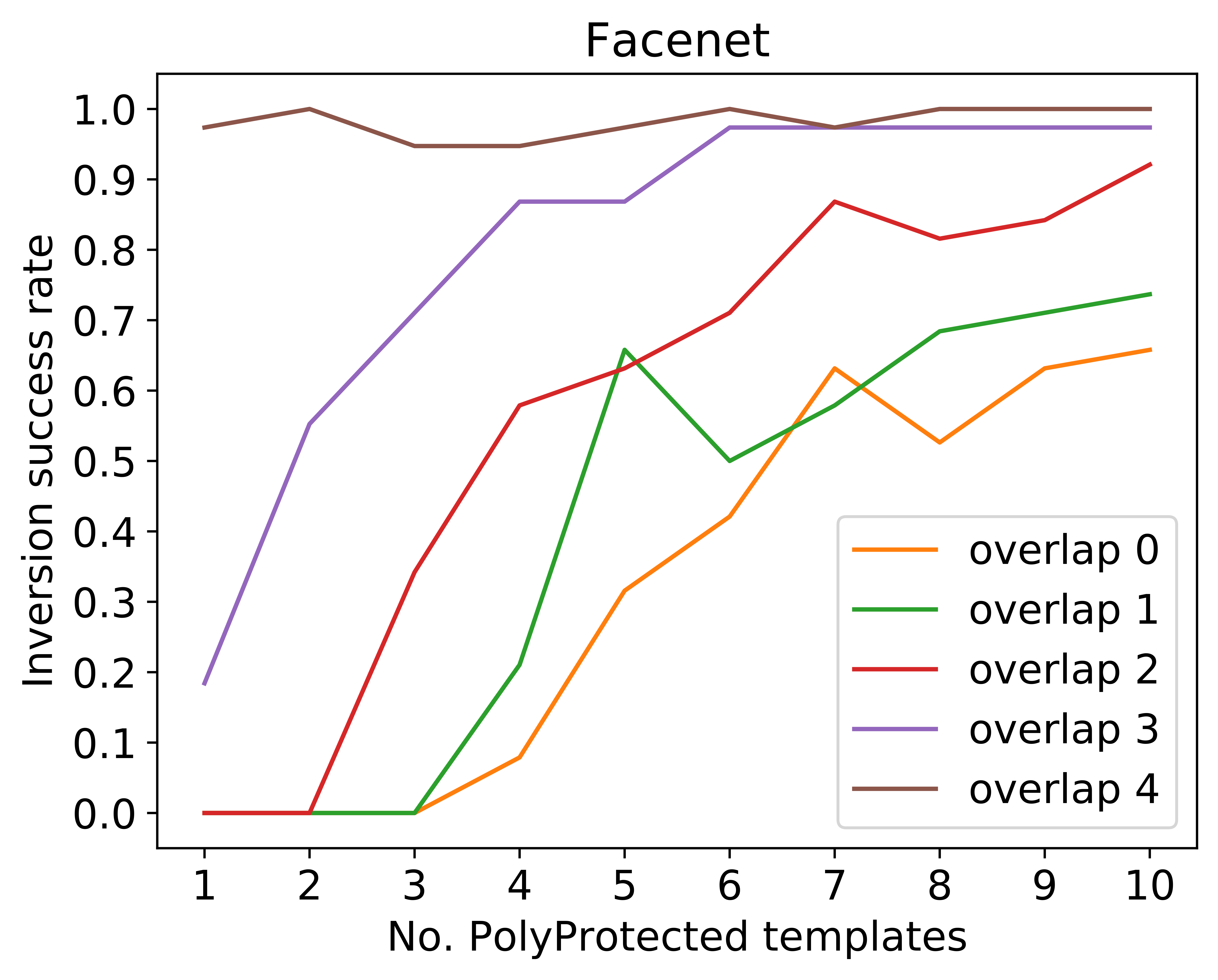}
\hfill
\includegraphics[width=0.49\columnwidth]{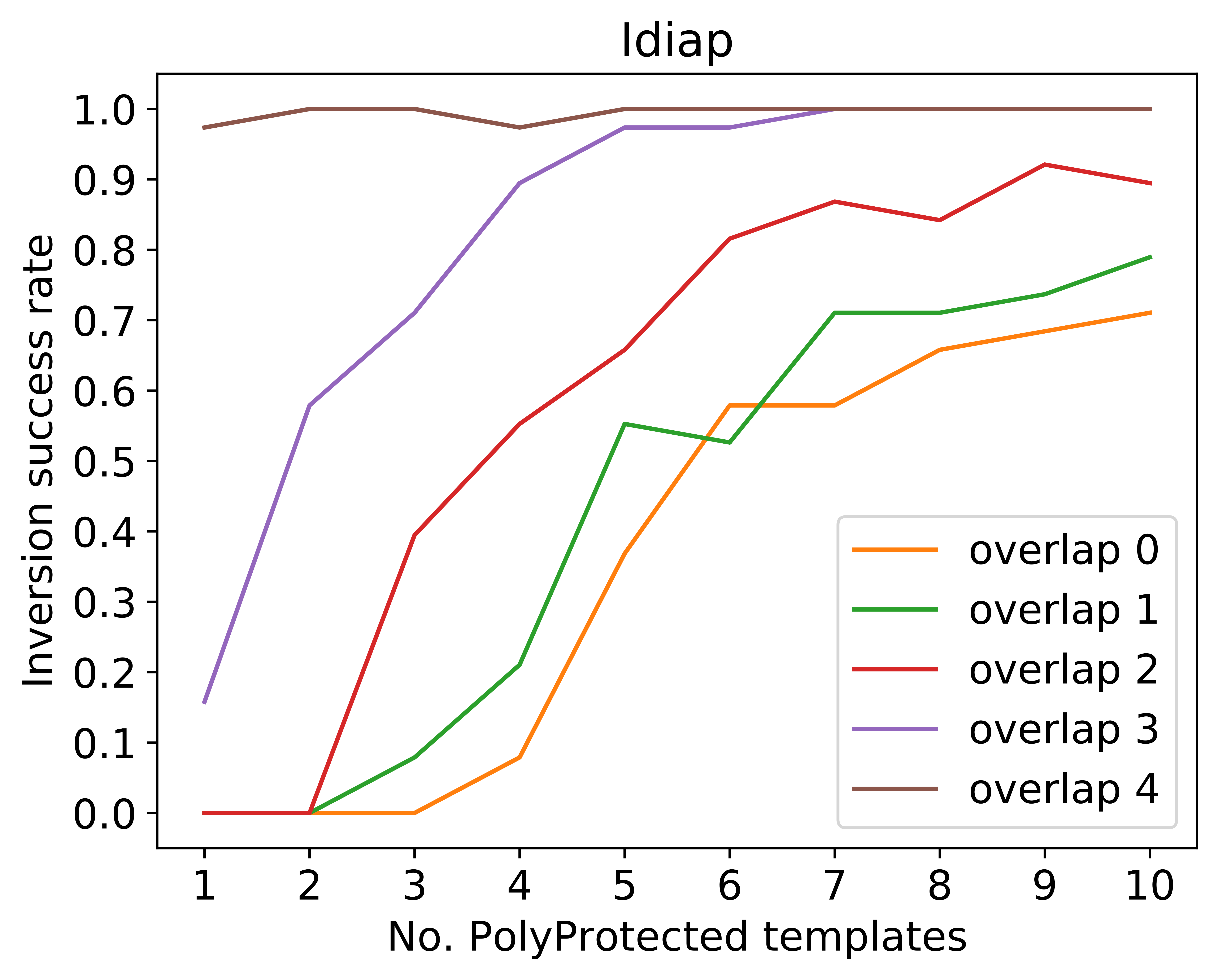}
\caption{Inversion success rate when an attacker is assumed to have access to multiple PolyProtected templates from the same face embedding. The results are based on the match threshold calculated at FMR = 0.1\% on the \textit{development} set of baseline face embeddings.}
\end{figure}

From Fig. 5, it is evident that, in general, the inversion success rate increases as the number of PolyProtected templates used in the attack increases. This makes sense, because access to a larger number of PolyProtected templates from the same embedding provides more information about that embedding, thereby allowing the numerical solver to approximate a more accurate solution to the $P \rightarrow V$ system of equations. As stated earlier, however, an overdetermined system of equations may result in an inconsistent solution set, thereby actually confusing the solver as to which solution is the correct one. This may explain the sudden drop in the inversion success rate in various curves in Fig. 5 (e.g., between 5 and 6 PolyProtected templates for overlap = 1 in both plots). So, a larger number of templates may not always translate to a higher inversion success rate in practice, even if this does appear to be the general trend.

Another general trend from Fig. 5 is that, the greater the amount of overlap used in the PolyProtect mapping, the fewer PolyProtected templates would be required to launch a Record Multiplicity Attack. This is in line with the irreversibility analysis in Section 4.3.2, which showed that the inversion success rate for a single $P$ increases as the amount of overlap increases. Since PolyProtected templates generated using a greater overlap contain more information about the original face embedding, it makes sense that the attacker would require more PolyProtected templates generated using a smaller overlap to achieve the same inversion success rate as for fewer PolyProtected templates generated using a larger overlap. So, we may conclude that, the smaller the amount of overlap used in the PolyProtect mapping, the less susceptible PolyProtect would be to a Record Multiplicity Attack in practice.

Having made these observations, we must emphasize that our ARM analysis assumed that all PolyProtected templates to which an attacker has access, have been generated from exactly the same face embedding, $V$. This should \textit{not} be the case in practice, even if the PolyProtected templates were acquired from the same system during re-enrollment of the same (compromised) user. Additionally, multiple PolyProtected templates belonging to the same subject enrolled in different applications may be generated by different systems (e.g., Facenet and Idiap), using different amounts of \textit{overlap} in the $V \rightarrow P$ mapping. So, our ARM analysis represents the best-case scenario for our simulated attacker, but in reality the susceptibility of PolyProtect to this type of attack should be much lower, especially as the attacker is unlikely to be \textit{fully informed}. 

Nevertheless, Fig. 5 indicates that even in this best-case scenario for the attacker, a high inversion success rate (e.g., $\approx$ 0.9 or above) for a commonly used match threshold is unlikely to be attainable for PolyProtected templates generated using overlaps of 0 or 1, even if the attacker had access to 10 templates from the same face embedding. Although such a high inversion success rate may be achievable for PolyProtected templates generated using an overlap of 2, the attacker would need to acquire a large number of templates from the same face embedding (i.e., 9 or 10), which may be considered practically infeasible\footnote{Hacking the database of even one system should be difficult, let alone 9 or 10 different systems, or the same system this many times.}. On the other hand, Fig. 5 suggests that a smaller number of PolyProtected templates would be needed when those templates have been generated using a larger overlap amount (i.e., 4-5 templates for an overlap of 3, or 1-2 templates for an overlap of 4). Although it is reasonable to assume that it would be difficult to acquire even 4-5 different PolyProtected templates from the same person in practice, 1-2 templates would be more feasible. So, as concluded for the single-template irreversibility analysis in Section 4.3.2, an overlap of 4 is not recommended in practice due to the high risk of inversion of the PolyProtected template(s) by a fully-informed attacker. In fact, in Section 4.3.2 we observed that an overlap of 2 generally seems to offer the best trade-off between irreversibility and recognition accuracy, while overlaps of 0 and 1 may be more suitable for lower-security applications, and an overlap of 3 could offer the best accuracy/irreversibility trade-off in higher-security applications. We make the same recommendation in light of our ARM analysis, while emphasizing once again that, if the priority is \textit{user privacy}, a smaller overlap should be employed in the PolyProtect mapping, as this would ensure the generation of \textit{more irreversible} PolyProtected templates.    

Finally, as for the irreversibility analysis in Section 4.3.2, it is currently impossible to present a meaningful, quantifiable comparison between PolyProtect's ARM susceptibility to that of the other face embedding protection schemes in the literature. This time, however, such a comparison is not only limited by the evaluation inconsistencies explained in Section 4.3.2, but also by the fact that this type of analysis is lacking for most of the proposed protection methods. For example, of the methods mentioned in Section 2, ARM susceptibility was only evaluated in \cite{m20}. In \cite{m20}, the authors state that the ARM attacking complexity is lower bounded by the effort of exhaustively searching for the user-specific key, thereby concluding that ARM does not reduce the complexity of inverting the protected templates. No analysis is presented to validate this claim from a practical point of view, however, especially considering the possibility of the user-specific key being leaked when it is decoded during an authentication attempt. So, it is difficult to draw a fair comparison between the susceptibility to ARM of the method in \cite{m20} and PolyProtect, in a practical context.  

\subsection{Unlinkability}

Assume that a certain PolyProtected face embedding, $P$, is enrolled in a face recognition system. This section investigates whether, in the event that $P$ is compromised (e.g., stolen from the system's database), we could \textit{renew} it, i.e., \textit{cancel} it and generate a replacement PolyProtected template, $P'$, by using different $C$ and $E$ parameters in the $V \rightarrow P$ mapping. The two templates, $P'$ and $P$, should be sufficiently different to ensure that they are \textit{unlinkable} (i.e., cannot be linked to the same identity). We also consider, therefore, the possibility of generating multiple \textit{diverse} PolyProtected templates from the same subject's face, for the purpose of enrolling this person in multiple applications without the risk of cross-matching their identity.

To conduct this evaluation in as realistic a setting as possible, we assumed that different PolyProtected templates belonging to the same subject would have been generated using different instances of that subject's face embedding. This is because the face embedding used to generate a particular PolyProtected template should be discarded during enrollment (i.e., only the PolyProtected template should be stored in the recognition system's database), so each new enrollment would require a new image of the subject's face, from which a new embedding would be generated. For example, let $V_1$ denote a subject's first face embedding, $P_1$ represent the corresponding PolyProtected template, and $C_1$ and $E_1$ denote the coefficients and exponents used to generate $P_1$, respectively. Now, assume that $P_1$ is compromised in some way, meaning that we must remove it from the database and replace it with a new PolyProtected template from the same subject's face. To achieve this, we ask the person to present a new sample of their face, from which the representative face embedding, $V_2$, is extracted. To protect $V_2$ via PolyProtect, we then generate new parameters, $C_2$ and $E_2$, which are used to create the new protected template, $P_2$. Alternatively, $P_1$ and $P_2$ could be used to enroll the same subject in two different applications. The following analysis considers whether $P_2$ is likely to be sufficiently different from $P_1$, such that they can effectively be seen as distinct, \textit{unlinkable} identities.

PolyProtect's unlinkability property was evaluated using the framework proposed in \cite{gb18}. We chose to adopt this framework, because it considers unlinkability from a \textit{practical} angle, which has been our focus in evaluating PolyProtect in this paper. Specifically, the method in \cite{gb18} measures unlinkability in the context of the \textit{mated} and \textit{non-mated} score distributions, which represent the comparison scores between different protected templates from the \textit{same} subject and between different protected templates from \textit{different} subjects, respectively. The unlinkability is measured in terms of two metrics: $D_{\leftrightarrow}(s)$, a local score-wise measure of the degree of linkability based on the likelihood ratio between mated and non-mated scores, and $D_{\leftrightarrow}^{\mathit{sys}}$, a global measure of the overall linkability of the underlying recognition system.

To evaluate the unlinkability of PolyProtected templates using the approach from \cite{gb18}, the first step was to select a number of different face embeddings from each subject in our adopted Mobio dataset, to simulate the enrollment of the same subject in multiple face recognition applications (or re-enrollment in the same application in the event that their protected template has been compromised). Then, the idea was to apply a \textit{different} set of $C$ and $E$ parameters to each face embedding, to generate its corresponding PolyProtected template, $P$. Based on the recommendation\footnote{At least 5 different protected templates per subject should be used. The authors used 10 in their experiments.} in \cite{gb18}, we randomly selected 10 different face embeddings per subject, resulting in 10 different PolyProtected templates per person. Then, each PolyProtected template was compared to every other PolyProtected template from the \textit{same subject} to generate a set of \textit{mated} comparison scores, and to all PolyProtected templates from \textit{every other subject} to generate a set of \textit{non-mated} comparison scores. This process was repeated for 10 trials, where in each trial a new set of 10 face embeddings was randomly selected for each subject in the dataset. The resulting 10 sets of mated and non-mated comparison scores were then concatenated (separately), and the concatenated scores were used to evaluate the unlinkability of the PolyProtected templates. For reference, we also calculated the unlinkability of the corresponding \textit{unprotected} embeddings in the same way. 

Fig. 6 shows the unlinkability plots\footnote{Produced using open-source code at: \url{https://bit.ly/3tYv3jw}} for our Facenet and Idiap PolyProtected and baseline (unprotected) systems, on Mobio's \textit{development} subset. Due to space restrictions, for the PolyProtected systems we show only the plots for overlap = 2, since this was the generally recommended value in Section 4.3, and Table 4 summarises the global $D_{\leftrightarrow}^{\mathit{sys}}$ measures for all overlaps.   

\begin{figure}[!h]
\centering
\includegraphics[width=0.48\columnwidth]{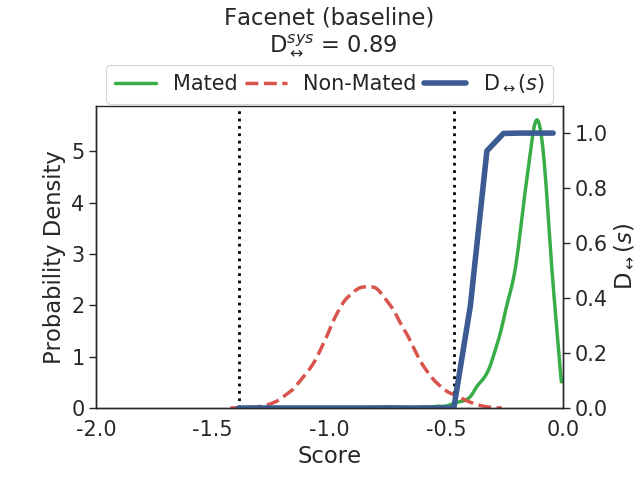}
\hfill
\includegraphics[width=0.48\columnwidth]{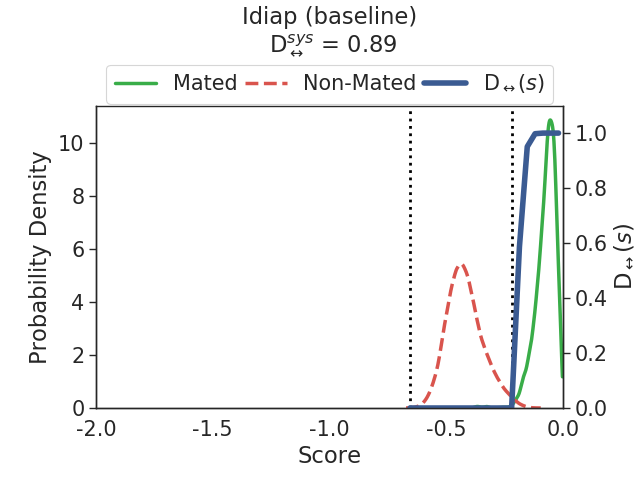}
\vfill
\includegraphics[width=0.48\columnwidth]{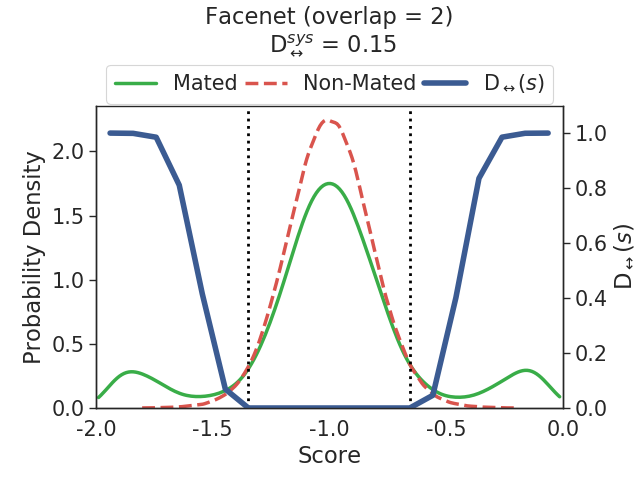}
\hfill
\includegraphics[width=0.48\columnwidth]{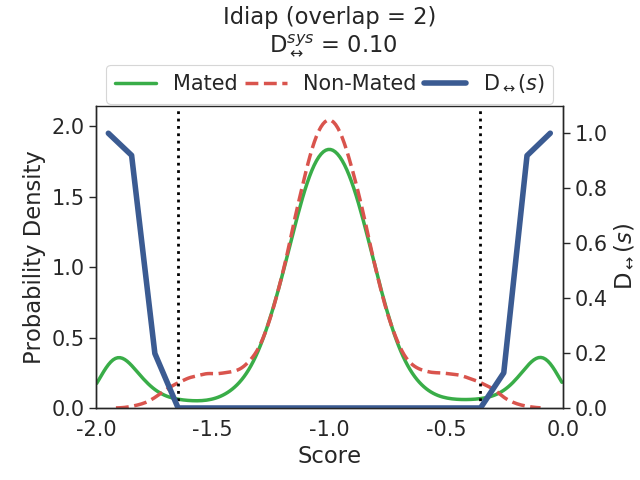}
\caption{Unlinkability plots for baseline and PolyProtected (overlap = 2) Facenet and Idiap systems, on Mobio's \textit{development} set. $D_{\leftrightarrow}(s)$ measures the score-wise degree of linkability, and the vertical dotted lines mark the score range within which linkability $\approx$ 0. $D_{\leftrightarrow}^{\mathit{sys}}$ (in the titles) indicates the overall system linkability, over the entire score range.}
\end{figure}

\begin{table}[!h]
\renewcommand{\arraystretch}{1.3}
\caption{$D_{\leftrightarrow}^{\mathit{sys}}$ for PolyProtected systems on Mobio's development set.}
\centering
\begin{tabular}{|l||c|c|c|c|c|}
\hline
\textbf{Overlap} & 0 & 1 & 2 & 3 & 4 \\
\hline
\textbf{Facenet} $\mathbf{D_{\leftrightarrow}^{\mathit{sys}}}$ & 0.14 & 0.15 & 0.15 & 0.15 & 0.16 \\
\hline
\textbf{Idiap} $\mathbf{D_{\leftrightarrow}^{\mathit{sys}}}$ & 0.09 & 0.10 & 0.10 & 0.11 & 0.12 \\
\hline
\end{tabular}
\end{table}

Note that we used the \textit{development} database subset instead of the \textit{evaluation} subset. This is because we wished to check whether \textit{full unlinkability} was attainable if the only requirement for the 10 different PolyProtected templates from the same subject was that their randomly generated $C$ and $E$ parameters were \textit{different}, but the extent of the differences between the resulting PolyProtected templates was not checked. In other words, our aim was to use the development subset as a sounding board for checking whether the aforementioned process would ensure the incorporation of sufficient \textit{diversity} into different PolyProtected templates from the same subject; if so, then we would apply the same procedure on the \textit{evaluation} Mobio subset, otherwise the approach would need to be reconsidered. 

There are several important observations from Fig. 6 and Table 4. Firstly, note that $D_{\leftrightarrow}^{\mathit{sys}}$ measures the overall system linkability, where a value of 0 would indicate that the system is fully \textit{unlinkable}, whereas a value of 1 would indicate that the system is fully \textit{linkable}. We observe that $D_{\leftrightarrow}^{\mathit{sys}}$ for our baseline systems, which use \textit{unprotected} face embeddings, is closer to 1, indicating that unprotected face embeddings from the same subject (e.g., used across different applications) are almost fully \textit{linkable}. On the contrary, the $D_{\leftrightarrow}^{\mathit{sys}}$ values for our PolyProtected systems are closer to 0, suggesting that different PolyProtected templates generated from the same subject's face embeddings are almost fully \textit{unlinkable}. 

Another observation is that $D_{\leftrightarrow}^{\mathit{sys}}$ increases slightly as the amount of overlap used in the PolyProtect mapping increases. This may be attributed to the fact that a greater overlap produces a higher-dimensional PolyProtected template, which contains more information from the original face embedding, so it may be a little easier to link higher-dimensional PolyProtected templates from the same identity. Having said this, the difference in $D_{\leftrightarrow}^{\mathit{sys}}$ between the different overlaps in Table 4 does not appear significant.  

Fig. 6 and Table 4 also show lower $D_{\leftrightarrow}^{\mathit{sys}}$ values for the PolyProtected Idiap system compared to the Facenet system, even though $D_{\leftrightarrow}^{\mathit{sys}}$ for the two baseline systems is the same. This is probably due to differences in the original embedding distributions, which are emphasized as a result of the PolyProtect mapping. Although the differences in the $D_{\leftrightarrow}^{\mathit{sys}}$ values across the two PolyProtected systems seem minor, this finding nevertheless indicates that the unlinkability of PolyProtected templates depends in part on the underlying face features, which makes sense in practice.

Another important observation\footnote{The same observation was made for overlaps of 0, 1, 3, and 4.} from Fig. 6 relates to the local score-wise measure of linkability, $D_{\leftrightarrow}(s)$. In both PolyProtected system plots, the value of $D_{\leftrightarrow}(s)$ within the score range marked by the vertical dotted lines is approximately 0. This indicates that, if we were to compare two PolyProtected templates and the resulting score was within that range, it would be almost impossible to link the two templates to the same identity; in other words, the systems are fully \textit{unlinkable} within that score range. Outside this score range, however, $D_{\leftrightarrow}(s)$ in both PolyProtected system plots gradually increases to 1. This implies that, if we were to compare two PolyProtected templates and the resulting score was found to lie towards the extremes of the score range, we could be almost certain that the two templates belong to the same identity; in other words, the systems tend towards full \textit{linkability} outside the score range marked by the vertical dotted lines.  

We believe that this phenomenon of full linkability at the extremes of the PolyProtected systems' score range (indicated by the two bumps in the \textit{mated} score distributions) is mainly due to the relationship between the \textit{signs} of the corresponding elements among each pair of PolyProtected templates that is compared to generate those scores. If the corresponding elements of the two PolyProtected templates all have the same signs (i.e., either both positive or both negative), then the resulting score will be close to 0.0. This is because the comparison score is a measure of the angle between the two template vectors, and the angle approaches 0 when the corresponding vector elements are similar in magnitude (since they come from the same subject's face) \textit{and} have the same signs (which is determined both by the signs of the original elements and the effects on the signs by the PolyProtect mapping). Alternatively, if the corresponding PolyProtected elements have opposite signs, the score will be close to -2.0. This effect is more prominent for PolyProtected templates from the \textit{same} subject's face embeddings, because these embeddings are much more similar than embeddings from different subjects' faces. Consequently, these similarities may be exaggerated to produce PolyProtected templates that are either as similar as possible (score $\approx$ 0.0) or as different as possible (score $\approx$ -2.0), depending on the choice of the $C$ and $E$ parameters used in the PolyProtect mappings.  

Based on the aforementioned observations, we conclude that our PolyProtected systems achieved a promising overall degree of unlinkability (indicated by the low $D_{\leftrightarrow}^{\mathit{sys}}$ values), but the systems were only fully unlinkable within a certain part of the score range. So, we investigated the possibility of achieving full unlinkability across the \textit{entire} score range, by selecting the $C$ and $E$ PolyProtect parameters in a stricter way, such that the \textit{mated} scores would be forced (as much as practically possible) to lie within the aforementioned range. In other words, we tested the possibility of removing the mated score distribution bumps at the extreme ends of the score range, through smarter PolyProtect parameter selection. To achieve this, we proceeded as follows. 

We established the full unlinkability score range on our \textit{development} dataset (as shown in Fig. 6), then applied that score range to the \textit{evaluation} dataset. Note that Mobio's development and evaluation datasets consist of \textit{different subjects}, so we can think of this approach as establishing the score range prior to system deployment on a dataset that simulates the application scenario, but when we do not know who will use the systems in practice. Since the full unlinkability score range was found to vary across the Facenet and Idiap face recognition systems and across the different overlaps, we used separate, system-specific and overlap-specific score ranges to select the $C$ and $E$ PolyProtect parameters that would be employed in generating the PolyProtected templates on Mobio's \textit{evaluation} dataset.

The same process that was applied on the \textit{development} subset to randomly select the face embeddings used in the unlinkability analysis (described earlier), was applied to the \textit{evaluation} dataset, and the same number of face embeddings and thus PolyProtected templates (i.e., 10 per subject, for each of 10 trials) was considered. The only difference was in the way that the $C$ and $E$ parameters were selected. For the \textit{development} dataset, we simply required the parameters to be \textit{different} for different PolyProtected templates for the same subject's face embeddings. For the \textit{evaluation} dataset, however, the parameter selection process was more strict.

Let $\{V_1, V_2, ..., V_{10}\}$ represent a set of 10 face embeddings randomly selected for a particular subject from Mobio's \textit{evaluation} dataset. For the unlinkability analysis, we needed to generate a PolyProtected template for each of these 10 embeddings, meaning that we had to generate 10 sets of $C$ and $E$ parameters. This was accomplished as follows. To produce $P_1$, which is the PolyProtected template of $V_1$, we simply generated the corresponding $C_1$ and $E_1$ randomly. Then, to produce $P_2$ from $V_2$, we also began by generating the corresponding $C_2$ and $E_2$ parameters randomly. At this point, ideally we would have liked to compare $P_2$ to $P_1$ and check if the resulting score was within the score range established for the corresponding face recognition system on Mobio's \textit{development} dataset; if this was the case, then $C_2$ and $E_2$ would be considered acceptable, otherwise a new set of parameters would be randomly generated until the aforementioned condition was satisfied. Unfortunately, however, such an approach would assume that the deployers of the face recognition system knew in advance all the PolyProtected templates that a particular person would use to enroll into different applications, which should not be the case in practice (but may be possible for template replacements within the \textit{same} system). What we \textit{can} assume, however, is that the \textit{parameters} used to generate those PolyProtected templates are known, because this is the only way to ensure that the same parameters are not used across different applications or for replacement PolyProtected templates within the same application.  

So, we adapted the process of selecting $C_2$ and $E_2$, such that, instead of being compared to $P_1$, the $P_2$ produced using these parameters was compared to the PolyProtected template produced from applying $C_1$ and $E_1$ to the \textit{same} face embedding, $V_2$. If the comparison score between the two PolyProtected templates was within the required score range, then $C_2$ and $E_2$ were accepted; otherwise, a new set of parameters was randomly generated until the aforementioned condition was satisfied. Once $P_2$ was successfully produced, we moved on to the selection of parameters $C_3$ and $E_3$, which would be used to produce $P_3$ from $V_3$. Similarly to the process used for $P_2$, we began by randomly generating $C_3$ and $E_3$ to produce $P_3$. This time, $P_3$ was compared to \textit{two} other PolyProtected templates: one resulting from applying $C_1$ and $E_1$ to $V_3$, and the other resulting from applying $C_2$ and $E_2$ to $V_3$. If \textit{both} comparison scores were within the required score range, $P_3$ was considered successful; otherwise, $C_3$ and $E_3$ kept being randomly generated until the score condition was satisfied. This process was continued until all 10 PolyProtected templates were successfully generated, where $P_{10}$ was compared to PolyProtected templates produced using $V_{10}$ and all 9 sets of previously-generated $C$ and $E$ parameters. 

The idea behind this strict process of selecting the $C$ and $E$ parameters was to ensure that \textit{different} PolyProtected templates generated from the \textit{same} face embedding would be \textit{unlinkable}. If we could achieve this, then it would be reasonable to assume that the selected parameters would also ensure unlinkability between PolyProtected templates generated from \textit{different face embeddings} belonging to the same subject (which should be the case if the face embeddings are quite similar). Note that this parameter selection process was conducted separately for each system (Facenet and Idiap) and each overlap in the set $\{0, 1, 2, 3, 4\}$, using the corresponding score ranges established on the \textit{development} dataset. Fig. 7 shows the unlinkability plots resulting from applying this \textit{strict} $C$ and $E$ parameter selection process on Mobio's \textit{evaluation} dataset. Due to space restrictions, Fig. 7 presents the unlinkability plots only for overlap = 2 for each of our two PolyProtected systems, and Table 5 summarises the $D_{\leftrightarrow}^{\mathit{sys}}$ values for all overlaps.  

\begin{figure}[!h]
\centering
\includegraphics[width=0.48\columnwidth]{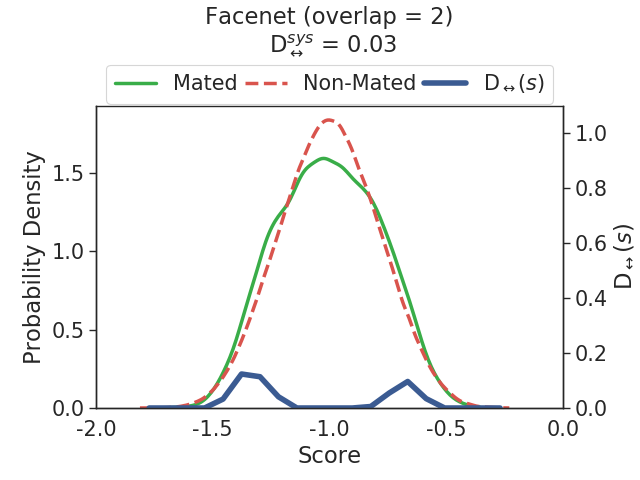}
\hfill
\includegraphics[width=0.48\columnwidth]{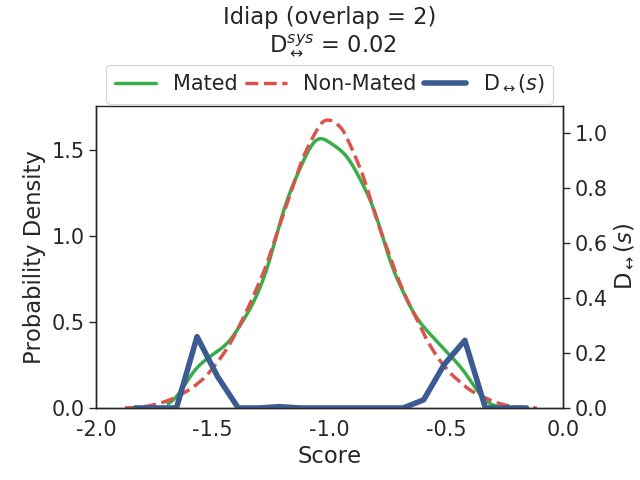}
\caption{Unlinkability plots for PolyProtected (overlap = 2) Facenet and Idiap systems, on Mobio's \textit{evaluation} set using strict $C$ and $E$ selection.}
\end{figure}

\begin{table}[!h]
\renewcommand{\arraystretch}{1.3}
\caption{$D_{\leftrightarrow}^{\mathit{sys}}$ for strict $C$ and $E$ selection on Mobio's evaluation set.}
\centering
\begin{tabular}{|l||c|c|c|c|c|}
\hline
\textbf{Overlap} & 0 & 1 & 2 & 3 & 4 \\
\hline
\textbf{Facenet} $\mathbf{D_{\leftrightarrow}^{\mathit{sys}}}$ & 0.03 & 0.04 & 0.03 & 0.04 & 0.03 \\
\hline
\textbf{Idiap} $\mathbf{D_{\leftrightarrow}^{\mathit{sys}}}$ & 0.03 & 0.02 & 0.02 & 0.01 & 0.01 \\
\hline
\end{tabular}
\end{table}

From Fig. 7 and Table 5, it is clear that $D_{\leftrightarrow}^{\mathit{sys}}$ for all PolyProtected systems is effectively 0. This suggests that, by employing a stricter parameter selection process for the PolyProtect mapping, it is possible to achieve almost full unlinkability for PolyProtected templates generated using all overlap amounts, on both the Facenet and Idiap systems. Furthermore, unlike in Fig. 6, $D_{\leftrightarrow}(s)$ in Fig. 7 does not approach anywhere near a value of 1 across the entire score range\footnote{The same trend was observed for all overlaps.}. A few scores are shown to have a non-zero $D_{\leftrightarrow}(s)$ measure, suggesting that there is a small probability of establishing a link between the PolyProtected templates being compared, if such a score is obtained; however, the low $D_{\leftrightarrow}(s)$ values even for these scores implies a very small degree of linkability, which is reflected in the near-zero $D_{\leftrightarrow}^{\mathit{sys}}$ measures. So, we may reasonably conclude that effectively full unlinkability of PolyProtected face embeddings is attainable in practice, provided that the $C$ and $E$ parameter generation is conducted in a manner smarter than naive random selection.

Note that, during our strict parameter selection process, we actually implemented a score range \textit{tolerance} if suitable $C$ and $E$ parameters could not be found within 100 tries. The tolerance started at 0 and was increased by 0.01 for every 100 failed parameter generation attempts. This was done to speed up the experiments, and we did not check for repetition across the 100 tries. So, we expect that more optimal unlinkability results could be obtained with a more stringent implementation, perhaps using a lower/zero tolerance. Nevertheless, the aim was to show that the unlinkability of our PolyProtected systems can be improved by adopting a smarter-than-random parameter selection method. The results in Fig. 7 and Table 5 prove that this is realisable in practice.

Of the example face embedding protection methods mentioned in Section 2, an unlinkability analysis was presented in \cite{m20, p21, l21}. These evaluations were also conducted using the framework proposed in \cite{gb18}, and the best $D_{\leftrightarrow}^{\mathit{sys}}$ values are approximately comparable to our results in Table 5. Since the full experimental procedure (e.g., the number of protected templates considered per subject) adopted for the unlinkability analysis of all these methods is not always clear, we cannot guarantee that our unlinkability comparison is perfectly fair. We may reasonably conclude, however, that in our respective evaluation scenarios, both PolyProtect and the protection methods in \cite{m20, p21, l21} were shown to satisfy the \textit{unlinkability} criterion to a high degree, using the same general evaluation framework.

\section{Conclusions and Future Work}

This paper proposed PolyProtect, a method for protecting face embeddings in neural-network-based face recognition systems. PolyProtect maps a face embedding to a lower-dimensional representation, via multivariate polynomials defined by user-specific coefficients and exponents. The recognition accuracy, irreversibility, and unlinkability of PolyProtect were evaluated on two open-source face recognition systems in a cooperative-user mobile face verification context, which represents PolyProtect's most likely application scenario in practice.

We showed that PolyProtect is capable of preserving or improving the baseline (unprotected) systems' recognition accuracy under normal operating conditions, and that the accuracy can be tuned by varying the amount of overlap used in the PolyProtect mapping. Acceptable accuracy is thus attainable even in the worst-case (albeit unlikely) scenario where all user-specific parameters are stolen.

Our irreversibility analysis, for a fully-informed attacker, simulated the feasibility of recovering an approximation of the face embedding from its PolyProtected template(s), using a numerical solver. The inversion success rate was calculated in terms of the comparison score between the returned solution and the true face embedding, which was assumed to be enrolled in the baseline (unprotected) face recognition system. At a commonly-used match threshold, PolyProtected templates generated using overlaps of 0-2 were found to be practically irreversible, those generated using an overlap of 3 partially reversible, and those generated using the maximum overlap of 4 almost fully reversible. As expected, the inversion success rate was demonstrated to decrease when the baseline face recognition system adopts a stricter match threshold, and increase when a more lenient threshold is used, though the difference was mainly noticeable for an overlap of 3. Access to multiple PolyProtected templates from the same face embedding was shown to increase the chances of a successful template inversion in the best-case scenario for the attacker, but in general a high success rate was found to be unattainable unless the maximum overlap was used in the PolyProtect mapping (not recommended) or the number of acquired templates was impractically large. We thus recommended carefully selecting the overlap parameter in practice, according to the desired irreversibility versus recognition accuracy trade-off.

Finally, an analysis of PolyProtect's unlinkability property showed that it is possible to achieve effectively full unlinkability between multiple PolyProtected templates from the same subject's face embeddings, particularly if the user-specific parameters employed in the PolyProtect mappings are selected in a smarter-than-random fashion. This suggests that it would be possible to generate sufficiently diverse PolyProtected templates from the same subject's face, such that: (i) compromised templates could be renewed (i.e., cancelled and safely replaced by new ones), and (ii) different templates could be enrolled in different applications without the risk of cross-matching.

Our focus in evaluating PolyProtect was on using \textit{practical} evaluation methodologies, to present insight into the method's robustness as a real-life face embedding protection scheme. The results indicate that the method is capable of satisfying the recognition accuracy, irreversibility, and unlinkability criteria, even under the toughest threat model that assumes a fully-informed attacker with complete knowledge of the system and all its parameters. We may thus reasonably conclude that PolyProtect shows promise in practice, which is important considering the urgent requirement for robust face embedding protection methods in real-life face recognition applications. 

Current plans for future work mainly include extending our evaluation to analyse the generalisability of PolyProtect across different types of face recognition systems (e.g., for embeddings extracted using different pre-trained face recognition models) and in diverse application scenarios (i.e., on datasets representing other face recognition contexts, in addition to cooperative-user mobile verification). In the same vein of thought, we intend to evaluate the suitability of PolyProtect for protecting embeddings extracted from \textit{other} biometric modalities (besides the face). Furthermore, we plan to work on expanding the unlinkability analysis to estimate the number of different PolyProtected templates that can be generated from the same biometric identity, consider additional types of practical attacks that can complement our existing irreversibility analysis, and broaden the recognition accuracy evaluation by investigating alternative (e.g., magnitude-specific) metrics for measuring the similarity between two PolyProtected templates.

\ifCLASSOPTIONcompsoc
  \section*{Acknowledgments}
\else
  \section*{Acknowledgment}
\fi

This material is based upon work supported by the Center for Identification Technology Research (CITeR) under Grant No. 20F-01I, and CITeR affiliates IDEMIA and SICPA.

\bibliographystyle{IEEEtran}
\bibliography{bibliography}


\vspace*{-8cm}

\begin{IEEEbiography}[{\includegraphics[width=1in,height=1.25in,clip,keepaspectratio]{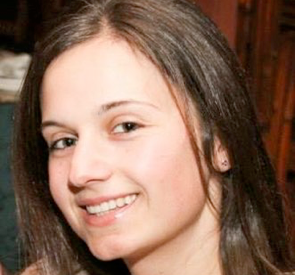}}]{Vedrana Krivoku\'ca Hahn} is a Research Associate at Idiap Research Institute (Switzerland). She received her PhD degree from the University of Auckland (New Zealand) in 2015. Vedrana works primarily on investigating biometric template protection methods and evaluation techniques. She has also been involved in consultation on the use and evaluation of face recognition systems in practice, and is currently contributing towards producing a digital, secure, and user-friendly personal data platform for European citizens. In addition to her research work, Vedrana teaches a full biometrics course as part of Idiap's Masters in Artificial Intelligence programme, and she is particularly interested in promoting ethical uses of biometric technologies.
\end{IEEEbiography}

\vspace*{-10cm}

\begin{IEEEbiography}[{\includegraphics[width=1in,height=1.25in,clip,keepaspectratio]{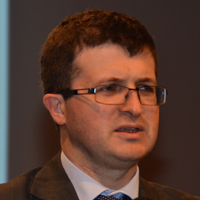}}]{S{\'e}bastien Marcel} heads the Biometrics Security and Privacy group at Idiap Research Institute (Switzerland) and conducts research on face recognition, speaker recognition, vein recognition, attack detection (presentation attacks, morphing attacks, deepfakes) and template protection. He received his Ph.D. degree in signal processing from Universit{\'e} de Rennes I in France (2000) at CNET, the research center of France Telecom (now Orange Labs). He is a lecturer at the Ecole Polytechnique F{\'e}d{\'e}rale de Lausanne and the University of Lausanne. He is also the Director of the Swiss Center for Biometrics Research and Testing, which conducts certifications of biometric products.
\end{IEEEbiography}




\end{document}